\documentclass[lettersize,journal]{IEEEtran}
\usepackage{amsmath,amsfonts}
\usepackage{algorithmic}
\usepackage{array}
\usepackage[caption=false,font=normalsize,labelfont=sf,textfont=sf]{subfig}
\usepackage{textcomp}
\usepackage{stfloats}
\usepackage{url}
\usepackage{verbatim}
\usepackage{pifont}
\usepackage{graphicx}
\usepackage{afterpage}
\usepackage{multirow}
\usepackage{amsmath}
\usepackage{makecell}
\usepackage{esint}
\usepackage{bigstrut}
\usepackage{amssymb}
\usepackage{amsmath}
\usepackage{multirow} 
\usepackage{ulem} 
\usepackage{cite}
\usepackage[table]{xcolor} 
\usepackage{colortbl} 

\usepackage[linesnumbered,ruled,vlined]{algorithm2e}
\def\BibTeX{{\rm B\kern-.05em{\sc i\kern-.025em b}\kern-.08em
    T\kern-.1667em\lower.7ex\hbox{E}\kern-.125emX}}
\usepackage{balance}

\begin{document}
\title{Spatial-frequency Dual-Domain Feature Fusion Network for Low-Light Remote Sensing Image Enhancement}
\author{Zishu Yao, Guodong Fan, Jinfu Fan, Min Gan, ~\IEEEmembership{Senior Member,~IEEE}, and C. L. Philip Chen, ~\IEEEmembership{Fellow,~IEEE}
\thanks{This work was supported in part by the National
	Natural Science Foundation of China under Grant 62073082, Grant 71701136,
	Grant U1813203, and Grant U1801262; in part by the Taishan Scholar
	Program of Shandong Province. (Zishu Yao and Guodong Fan contributed equally to this work.) (Corresponding
	author: Jinfu Fan.)}
\thanks{Zishu Yao, Guodong Fan, Jinfu Fan and Min Gan are with the College of Computer Science \& Technology, Qingdao University, Qingdao 266071, China (e-mail: yaozishu@qdu.edu.cn; fgd96@outlook.com; aganmin@aliyun.com; fan\_jinfu@163.com).}
\thanks{C. L. Philip Chen is with the College of Computer Science \& Technology, Qingdao University, Qingdao 266071, China and the School of Computer Science \& Engineering, South China University of Technology, Guangzhou 510641, China (e-mail: philip.chen@ieee.org).}
}

\markboth{Journal of \LaTeX\ Class Files,~Vol.~18, No.~9, September~2020}%
{How to Use the IEEEtran \LaTeX \ Templates}

\maketitle

\begin{abstract}

Low-light remote sensing images generally feature high resolution and high spatial complexity, with continuously distributed surface features in space. This continuity in scenes leads to extensive long-range correlations in spatial domains within remote sensing images. Convolutional Neural Networks, which rely on local correlations for long-distance modeling, struggle to establish long-range correlations in such images. On the other hand, transformer-based methods that focus on global information face high computational complexities when processing high-resolution remote sensing images. From another perspective, Fourier transform can compute global information without introducing a large number of parameters, enabling the network to more efficiently capture the overall image structure and establish long-range correlations. Therefore, we propose a Dual-Domain Feature Fusion Network (DFFN) for low-light remote sensing image enhancement. Specifically, this challenging task of low-light enhancement is divided into two more manageable sub-tasks: the first phase learns amplitude information to restore image brightness, and the second phase learns phase information to refine details. To facilitate information exchange between the two phases, we designed an information fusion affine block that combines data from different phases and scales. Additionally, we have constructed two dark light remote sensing datasets to address the current lack of datasets in dark light remote sensing image enhancement. Extensive evaluations show that our method outperforms existing state-of-the-art methods. The code is available at https://github.com/iijjlk/DFFN.

\end{abstract}

\section{Introduction}
\label{introduction}

Remote sensing (RS) technology aims to capture clear ground scene information. Compared to ordinary images, remote sensing images display richer color information and broader spatial information. However, due to nighttime conditions and other inherent natural factors, images captured in low-light environments often suffer from color degradation and loss of detail. This inevitably affects many downstream tasks\cite{10175627,zhang2021rethinking} that rely on remote sensing images, such as disaster recovery, wildlife monitoring, and environmental protection. Therefore, the development of algorithms dedicated to enhancing low-light remote sensing images is crucial.


Compared to images in traditional Low-light image enhancement tasks\cite{9743313,li2021low,zhang2021better,10219916,fan2022multiscale,ma2021learning,li2023low}, RS images have larger lengths and widths. Some surface features are often continuously distributed in space, such as forests covering several square kilometers or roads traversing the entire image. The continuity of such scenes results in remote sensing images exhibiting long-distance correlations in spatial aggregation. Convolutional Neural Networks (CNNs)\cite{9812717,zhou2023underwater,10234460} rely on local perception. For high-resolution images, the receptive field of a single convolutional operation struggles to capture larger spatial structures or global information in the image.
\begin{figure}[!t]
	\centering
	\includegraphics[width=0.95\columnwidth]{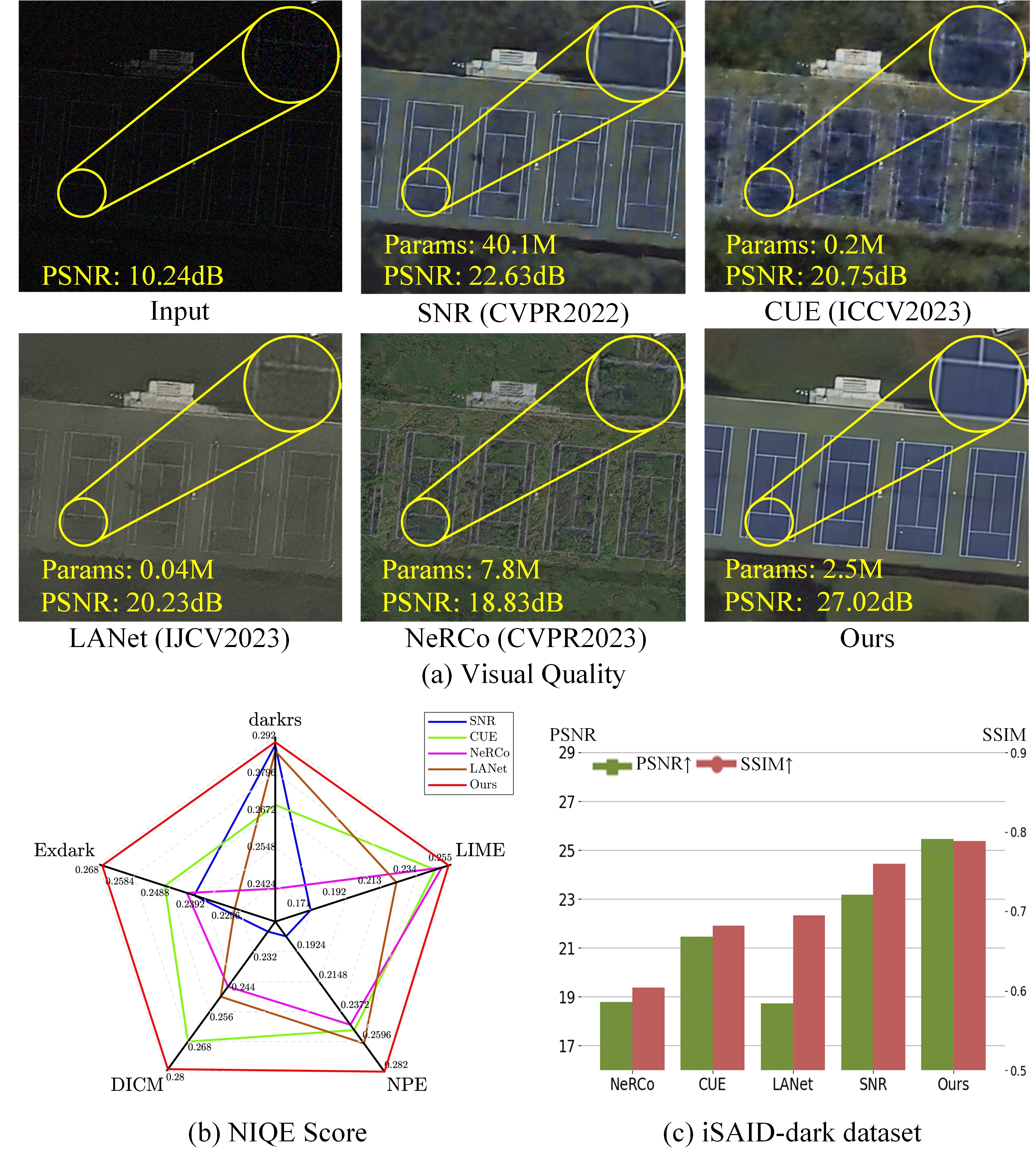}
	\caption{Comparison between the latest state-of-the-art methods and our approach. As shown in the zoomed-in region, these methods exhibit color distortion and excessive noise, leading to reduced visual quality. In contrast, our method presents vibrant colors and sharp outlines. Transformer-based method SNR\cite{xu2022snr} achieves similar visual effects to ours but with 40.1M parameters. Additionally, in (b), we compute the NIQE\cite{mittal2012making} scores on a no-reference dataset and display them in inverse form. Furthermore, in (c), scores of SSIM\cite{wang2004image} and PSNR\cite{huynh2008scope} metrics on a reference dataset are shown. It can be easily observed that our method significantly outperforms others.}
	\label{fig1}
\end{figure}
Although deep convolutions can expand the receptive field, as the network depth increases, the loss of detailed information gradually occurs, making it challenging for models to establish long correlations in RS images. Recently, a number of transformer methods\cite{9992208,10196016,10463068,10167502,10382425}, with Vision Transformer\cite{dosovitskiy2020image} as a prominent example, have successfully addressed the challenge of establishing long-range dependencies by implementation of self-attention mechanisms. However, the quadratic complexity problem remains difficult to solve. Additionally, these methods tend to overfit when trained on small datasets, significantly reducing their reliability in practical applications. Fig. \ref{fig1} (a) presents visual results and number of parameters for several representative methods. It can be observed that SNR, built upon the transformer, achieves suboptimal visual results, yet concurrently bears an onerous computational burden.

\begin{figure}[h]
	\centering
	\includegraphics[width=1.0\columnwidth]{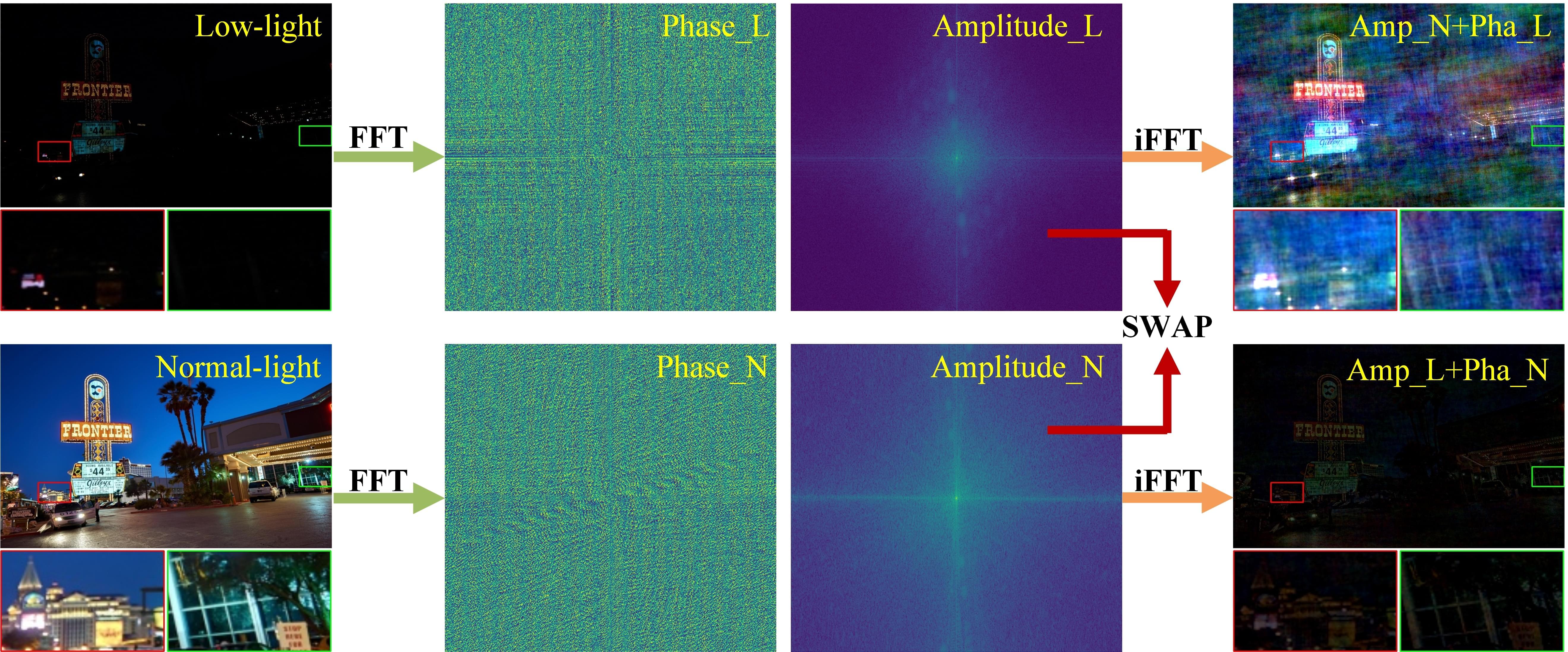}
	\caption{Our Motivation. As observed in the figure, when swapping the amplitude and phase of low-light images with those of normal images, the overall ambient light of the image containing degraded amplitude darkens while maintaining clear details. On the other hand, in the image containing phase degradation, there is no change in brightness, but the image details vanish. This indicates that Fourier transform can decouple degradation information.}
	\label{fig2}
\end{figure}
Most existing methods\cite{10298274,10177702,ma2023bilevel} process images in the spatial domain based on pixel perspectives. For instance, Xing et al.\cite{xing2023clegan} proposed an unsupervised GAN framework for low-light remote sensing image enhancement based on the self-similarity characteristics of remote sensing images. Singh et al.\cite{singh2022low} designed a multi-resolution Rnet that preserves high-resolution information while retaining low-resolution standards through different data streams. In contrast, we shift our focus to the Fourier domain, motivated by two unique phenomena inherent in the Fourier domain: 1) Under Fourier transformation, global information of the image can be extracted without introducing a large number of parameters\cite{wang2023fourllie}. 2) It is well-known that Fourier phase preserves high-level semantics, while amplitude contains low-level features\cite{oppenheim1979phase}. Fig. \ref{fig2} displays the visualization image of exchanging the Fourier amplitude and phase between normal and low-light images. It can be observed that brightness degradation exists in the amplitude, while other information is preserved in the phase. This phenomenon suggests that the Fourier transform can decouple the degradation of low-light images, breaking down the complex coupling problem into two relatively easier-to-solve sub-problems. Existing methods have explored the frequency domain information, but still present many shortcomings. For example, Wang et al.\cite{wang2023fourllie} opts to simultaneously recover both amplitude and phase information without considering spatial information. Li et al.\cite{li2023embedding} only performs Fourier decomposition on images within the network, yet the loss function continues to make comparisons with the ground truth (GT) solely in the spatial domain, lacking effective supervision. This undoubtedly makes the recovery process more difficult. Therefore, the challenge of how to integrate frequency domain information with spatial features and efficiently apply this integration for the enhancement of remote sensing low-light images remains unresolved.

Additionally, multi-stage architecture networks\cite{zhu2020eemefn,li2021low,wang2023fourllie,mou2022deep} often face issues with insufficient information interaction. Most existing works directly use the output of the previous stage as the input for the next stage. However, such methods do not consider the rich information contained in the intermediate features, which can easily lead to information loss. Other works adopt a one-to-one correspondence approach to pass information from the previous stage to the next stage, effectively compensating for the information loss and detail blurring caused by the upsampling and downsampling processes. However, these solutions still only consider the interaction of information at the same scale, neglecting the interaction of cross-scale features within and between stages, making it difficult to obtain more refined contextual features.

In this work, based on two unique phenomena of the Fourier transform, we propose a Dual-domain Feature Fusion Network (DFFN). Specifically, the DFFN consists of two-stage transformations to progressively restore high-quality images. Initially, the amplitude illumination phase learns the amplitude mapping from low-light to normal lighting conditions to brighten the image. Subsequently, the phase refinement phase enhances image details further. To integrate frequency domain and spatial features, we designed the Dual-Domain Amplitude Block (DDAB) for the amplitude illumination phase and the Dual-Domain Phase Block (DDPB) for the phase refinement phase. By splitting the Low-light Remote Image Enhancement task into two sub-problems, we reduce the complexity of degradation coupling. 

Meanwhile, to improve information interaction capability between the two-stage networks, we designed an Information Fusion Affine Module (IFAM) that integrates rich feature information from different stages and scales, achieving adaptive learning through dynamic  fusion with affine filtering. The obtained feature information is convolved with the decoder features of the phase refinement stage to enhance the network representation capability.
   
\begin{figure*}[!t]
	\centering
	\includegraphics[width=0.7\textheight]{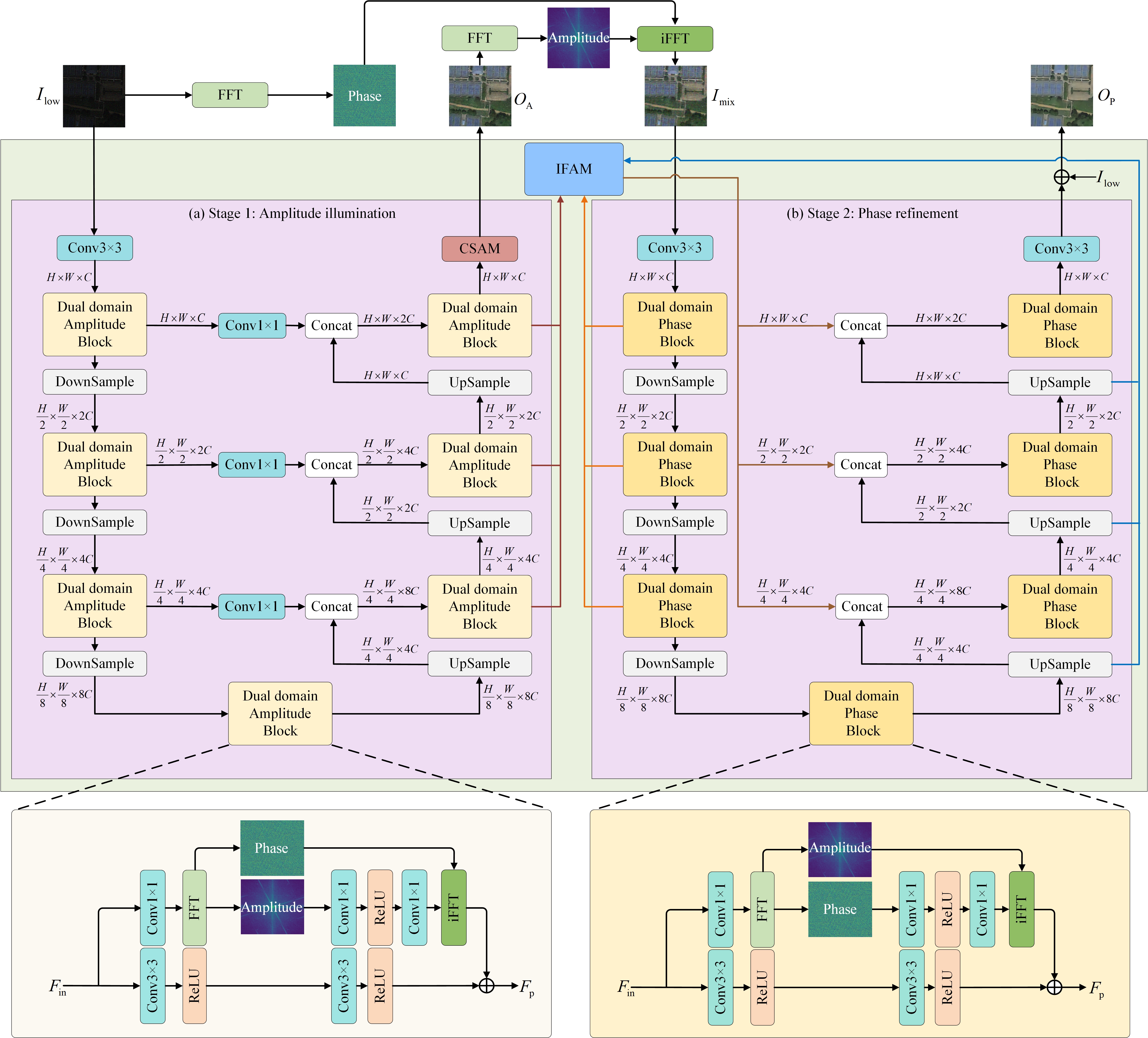}
	\caption{Overall network structure diagram. The specific structures of DDAB and DDPB are displayed below the diagram. The low-light image $I_{low}$ first undergoes an amplitude illumination stage to obtain a preliminary enhanced image $O_A$, followed by a phase refinement stage to obtain the final enhanced image $O_P$. In this process, the feature information from different stages is passed into IFAM for mapping to enhance the contextual representation capability of the model.}
	\label{fig3}
\end{figure*}


In addition, due to the lack of available image pairs, the development of existing remote sensing low-light enhancement tasks has been slow. In this paper, we constructed two large datasets, with the paired dataset iSAID-dark used for training and the real dataset darkrs used for testing.

In summary, our work contributes the following:

\begin{itemize}

    \item{We propose a Dual-domain Feature Fusion Network for low-light remote sensing image enhancement, which decomposes the degradation problem into two relatively easier-to-solve sub-problems through a two-stage network.}
    
    \item{We design an Information Fusion Affine Module to facilitate information interaction across scales, stages, and domains within the two-stage network. This module promotes information fusion between the two stages and enhances the representation of global context.}

    \item{We construct two low-light remote sensing datasets, including a real dataset captured by UAVs at night and a large-scale synthetic dataset. These datasets provide paired data for low-light remote sensing image enhancement and real low-light scenes for testing purposes.}

    \item{We conduct extensive experiments on multiple public benchmarks for comprehensive evaluation, demonstrating that our method performs well compared to state-of-the-art methods. Additionally, our approach shows excellent trade-offs between model complexity and performance.}
    
\end{itemize}

\section{related work}
\subsection{Low-light image enhancement}
In recent years, Low-light image enhancement has developed rapidly, which can mainly be classified into three categories: methods based on value mapping, model-based methods, and data-driven methods.
	
Value mapping-based methods enhance image exposure by altering the pixel distribution through mapping relationships. Notably, such methods include histogram equalization \cite{arici2009histogram,celik2011contextual,ibrahim2007brightness} and curve transformations, such as gamma transformation \cite{huang2012efficient,singh2017novel}. However, due to the lack of consideration for pixel context information, these methods are prone to causing local illumination inconsistencies and color abnormalities. 

Model-based methods \cite{fu2016weighted,jobson1997multiscale,xu2024seeing}, grounded in Retinex theory, view images as the product of reflectance and irradiance. By dividing low-light images by the irradiance image, the resulting reflectance is used as the final enhanced image. For example, Li et al.\cite{li2018structure}, building upon traditional Retinex theory and taking noise maps into account, successfully developed a noise-robust Retinex model. Guo et al.\cite{guo2016lime} estimated the illumination for each pixel using the maximum channel prior and added a prior structure on top of it to refine the illumination. The crux of such methods lies in designing well-crafted regularization functions. However, these regularization functions are based on prior assumptions, which, when not valid, greatly diminish the effectiveness of these methods.

Data-driven approaches \cite{shen2023mutual,liang2022self,10147801} are based on large-scale datasets and enhance network performance through the design of various complex model structures. For instance, Ma et al.\cite{ma2022learning} proposed a context-sensitive decomposition network that focuses on the contextual dependencies in the feature space, estimating lighting and reflections at the feature level to ultimately achieve high-quality enhancement effects. Li et al.\cite{li2022learning} designed a pixel-level curve estimation method and developed a series of loss functions that, while ensuring high-speed computation, yielded excellent results. Wang and colleagues\cite{wang2023low} combined gamma correction with transformers, establishing dependencies for all pixels from local to global, efficiently recovering dark areas. 

Although existing low-light image enhancement methods have achieved remarkable enhancement effects, due to the lack of corresponding paired datasets and the rarity of methods specifically designed for remote sensing images, current mainstream methods struggle to achieve ideal results in the task of enhancing dark light remote sensing images. Recently, Ling et al. \cite{10136217} proposed an unpaired training method for low-light remote sensing image enhancement, which maximizes the mutual information between dark and normal light images in a generative adversarial network through self-similarity based contrastive learning. However, due to limited supervision, the final enhancement effect is not satisfactory. To facilitate the development of this task, we propose a large-scale paired dataset for dark light remote sensing image enhancement to provide effective supervision.

\begin{figure*}[!t]
	\centering
	\includegraphics[width=2.0\columnwidth]{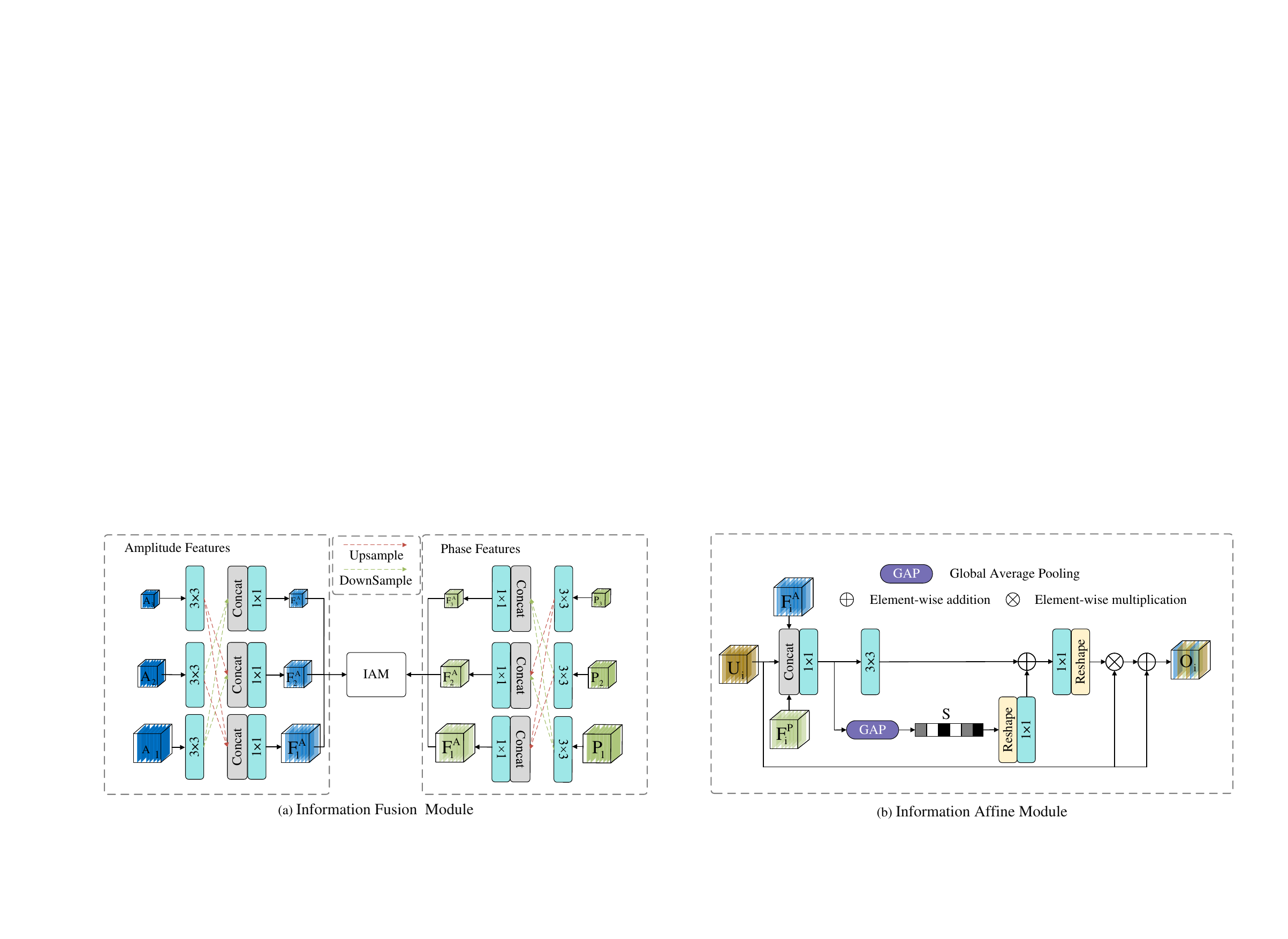}
	\caption{The specific structure of the proposed IFAM consists of two key components: the Information Fusion Module and the Information Mapping Module. $U_{i}$ represents the feature obtained after the $i$-th upsampling in the Phase refinement stage.}
	\label{fig4}
\end{figure*}

\subsection{Multi-stage Network}
 The multi-stage network is designed to decompose challenging tasks into several easier sub-tasks and solve them progressively to achieve optimal results. However, current multi-stage networks still have some issues. Specifically, existing multi-stage networks can be divided into the following two categories: 1) "result fusion" and 2) "cross-level fusion".

	The first category often uses the output of the previous stage as the input for the next stage. For example, Zhu et al. \cite{zhu2020eemefn} proposed an Edge-Enhanced Multi-Exposure Fusion Network (EEMEFN), where they initially brighten dark images and then enhance the edges of the preliminary enhanced results to obtain the final restored outcome. Li et al. \cite{li2021low} utilized a recurrent network to learn low-light image enhancement, with each stage taking the attention map and the previous stage's output as input, then performing recursive learning through the same residual module. Wang et al. \cite{wang2023fourllie}, in their proposed dual-stage network, first enhanced image brightness by estimating magnitude information, followed by the introduction of a signal-to-noise ratio (SNR) map to restore image details. Although these methods achieve good visual effects, they all neglect the rich intermediate features, and solely using the preliminarily enhanced image can lead to information loss, resulting in blurred details.

	The second category aims to integrate rich contextual features across multiple stages, alleviating information loss and gradient vanishing issues in deep networks. For instance, in the proposed Deep Generalized Unfolding Network (DGUNet), Mou et al. \cite{mou2022deep} passed the decoder features from each stage to the corresponding module in the next stage to capture rich informational features. However, these methods typically only consider the interactions between paired features and do not account for feature interactions across different scales.

In contrast, our proposed IFAM effectively combines feature information across different stages and scales. By utilizing dynamic fusion and affine filtering for adaptive learning, it significantly enhances the network's representational capacity.

\subsection{Fourier Transform in Neural Networks}
Recently, due to the unique properties of images in the Fourier space, such as capturing global representations, Fourier frequency information \cite{wang2023fourllie,li2023embedding,huang2022deep} has attracted widespread attention. A series of works have been based on Fourier domain transformations to enhance the generalization performance of networks. For instance, Mao et al. \cite{mao2021deep} designed a residual Fourier block to address issues like the potential neglect of low-frequency information and the difficulty in modeling long-distance information in Resblocks. Roman et al. \cite{suvorov2022resolution} used fast Fourier convolution to design a large mask inpainting network, achieving impressive results even in challenging scenarios. Dario et al. \cite{fuoli2021fourier} employed Fourier loss to aid networks in recovering high-frequency information in super-resolution images. Meanwhile, the field of remote sensing has seen extensive exploration of frequency domain information \cite{zheng2022hfa}. For example, Zhou et al. \cite{zhou2022spatial} demonstrated excellent performance in the pan-sharpening task by exploring the relationship between Pan images and low-resolution multispectral (MS) images with frequency domain information. Yu et al. \cite{yu2022snnfd} proposed a spiking neural network in the frequency domain (SNNFD), which enhances the model's transferability and feature extraction capability for buildings of different sizes by integrating frequency and spatial domain learning. Inspired by these works, we proposed DFFN based on Fourier transform, which effectively decouples low-light degradation and achieves superior enhancement results.

\section{Method}

In this section, we first introduced the Fourier information transformation. Subsequently, we described the overall architecture of the DFFN and the basic units constituting the network. Finally, we elaborated on the IFAM and the loss functions used in the DFFN.

\begin{figure}[h]
	\centering
	\includegraphics[width=0.95\columnwidth]{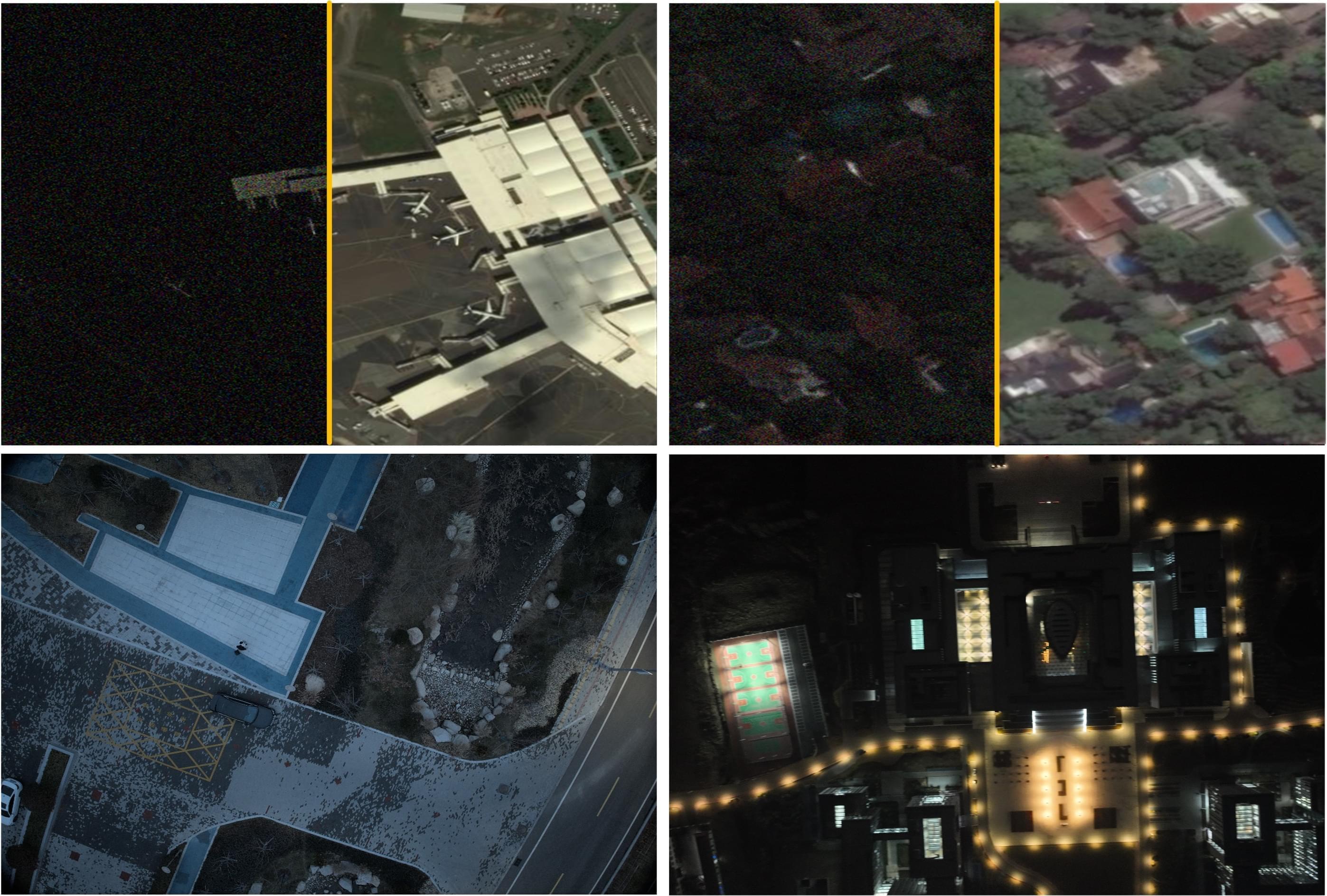}
	\caption{Samples from the proposed iSAID-dark(Up) and darkrs(Down) dataset.}
	\label{fig5}
\end{figure} 
  
\subsection{Fourier Frequency Information}
In this subsection, we will briefly introduce Fourier frequency information. Specifically, consider an image 
$I\in  \mathbb{R}^{H \times W \times 1 }$, whose Fourier transformation function  $\mathcal{F}$ can be represented as:
\begin{align}
    \mathcal{F}(X)(i,j) = \frac{1}{\sqrt{HW}} \sum_{h = 0}^{H-1}\sum_{w = 0}^{W-1}X(h,w)e^{-j2\pi (\frac{h}{H}i + \frac{w}{W}j)}, 
\end{align}
where $k$ represents the imaginary unit, $i$ and $j$ represent the horizontal and vertical coordinates respectively, $h$ and $w$ represent spatial coordinates, and the inverse process of $\mathcal{F}$ is denoted as $\mathcal{F}^{-1}$. 

Correspondingly, the amplitude $\mathcal{A}(I)$ and phase $\mathcal{P}(I)$ information can be defined as:
\begin{equation}
\left\{
\begin{aligned}
    &\mathcal{A}(I)(i,j) = \sqrt{R^{2}(X)(i,j)+I^{2}(X)(i,j)}, \\
    &\mathcal{P}(I)(i,j) = \arctan\left [ \frac{I(X)(i,j)}{R(X)(i,j)} \right],
\end{aligned}
\right.
\end{equation}
where $I(X)(i,j)$ and $R(X)(i,j)$ denote the imaginary and real component of $F(X)(i,j)$. For feature maps or color images, Fourier calculations are computed separately for each channel. The $R(X)(i,j)$ and $I(X)(i,j)$ can also be obtained by:
\begin{equation}
\left\{
\begin{aligned}
    &R(X)(i,j) = \mathcal{A}(X)(i,j) \times \cos(\mathcal{P}(X)(i,j)), \\
    &I(X)(i,j) = \mathcal{A}(X)(i,j) \times \sin(\mathcal{P}(X)(i,j)).
\end{aligned}
\right.
\end{equation}

\subsection{Network Architecture}

The structure of DFFN is shown in Fig. \ref{fig3}. Specifically, in the amplitude illumination stage, image brightness is restored by adjusting the amplitude, while in the phase refinement stage, phase information is learned to refine fine details. Since the two stages have different subtasks, their input information and supervision targets should also differ. The low-light image $I_{low}$ as the input for the first stage. The supervision target is defined as $\mathcal{F}^{-1}(\mathcal{A}(I_{gt}),\mathcal{P}(I_{low}))$. To prevent the first stage network from affecting the phase information, the output $O_{A}$ is not directly used as the input for the second stage. Instead, let $I_{mix}$ = $\mathcal{F}^{-1}(\mathcal{A}(O_{A}),\mathcal{P}(I_{low}))$, and the supervision target is $I_{gt}$.

Although our primary motivation lies within the Fourier domain, the utilization of spatial branches is necessary. This is because spatial branches and Fourier branches complement each other. Spatial branches, employing convolution operations, effectively model structural dependencies in the spatial domain. Fourier branches can engage in global information, facilitating the unraveling of energy and degradation. Therefore, for more efficient information recovery, we introduce DDAB and DDPB.  

    \subsubsection{Dual-Domain Amplitude Block}
    {
        As shown in the bottom-left of Fig. \ref{fig3}, The DDAB consists of two parallel branches: spatial domain branch and frequency domain branch. The spatial domain branch has two 3$\times$3 Conv layers to capture spatial information. In the frequency domain branch, the feature stream $F_{in}$ first undergoes 1$\times$1 Conv to refine the features. Then, FFT is applied to obtain the amplitude component  $\mathcal{A}(F_{in})$ and the phase component  $\mathcal{P}(F_{in})$. The amplitude component $\mathcal{A}(F_{in})$ is passed through a Conv layer with two 1 $\times$ 1 kernels (to avoid damaging the frequency domain structure, only 1 $\times$ 1 Conv is used here). Meanwhile, the corresponding phase information $\mathcal{P}(F_{in})$ is preserved through a separate branch. Subsequently, we use the iFFT to map $\mathcal{A}(F_{in})$ and $\mathcal{P}(F_{in})$ back to the image space to obtain frequency domain features. Finally, we integrate the spatial and frequency domain features through residual connections. The process can be described as follows:
        \begin{equation}
        \begin{aligned}
        \left\{ \begin{array}{l}
        F_{sa} = \sigma_{3\times3}\left[ \sigma_{3\times3}\left( F_{in} \right) \right], \\
        \mathcal{A}(F_{in}), \mathcal{P}(F_{in}) = \mathcal{F}[\Theta_{1\times1}\left( F_{in} \right)], \\
        F'_{A} = \mathcal{F}^{-1}\left\langle \Theta_{1\times1}\left\{ \sigma_{1\times1} \left[ \mathcal{A}(F_{in}) \right] \right\}, \mathcal{P}(F_{in}) \right\rangle, \\
        F_{A} = F_{sa} + F'_{A},
        \end{array} \right.
        \end{aligned}
        \end{equation}
        where $\sigma(\cdot )$ represents Conv and ReLU operation, $\mho(\cdot)$ represents concatenation operation, and $\Theta(\cdot)$ represents Conv operation.
    }  
    
    \subsubsection{Dual-Domain Phase Block}
    {
        Compared to DDAB, DDPB simply swaps the operations of amplitude and phase. It can be described as:
            \begin{equation}
            \begin{split}
               F^{'}_{P} = \mathcal{F}^{-1}\left \langle \Theta _{1\times1}\left \{ \sigma _{1\times1} \left [ \mathcal{P}(F_{in}) \right ]  \right \} , \mathcal{A}(F_{in}) \right \rangle . 
            \end{split}
            \end{equation}
    }

\begin{figure}[!h]
	\centering
	\includegraphics[width=0.97\columnwidth]{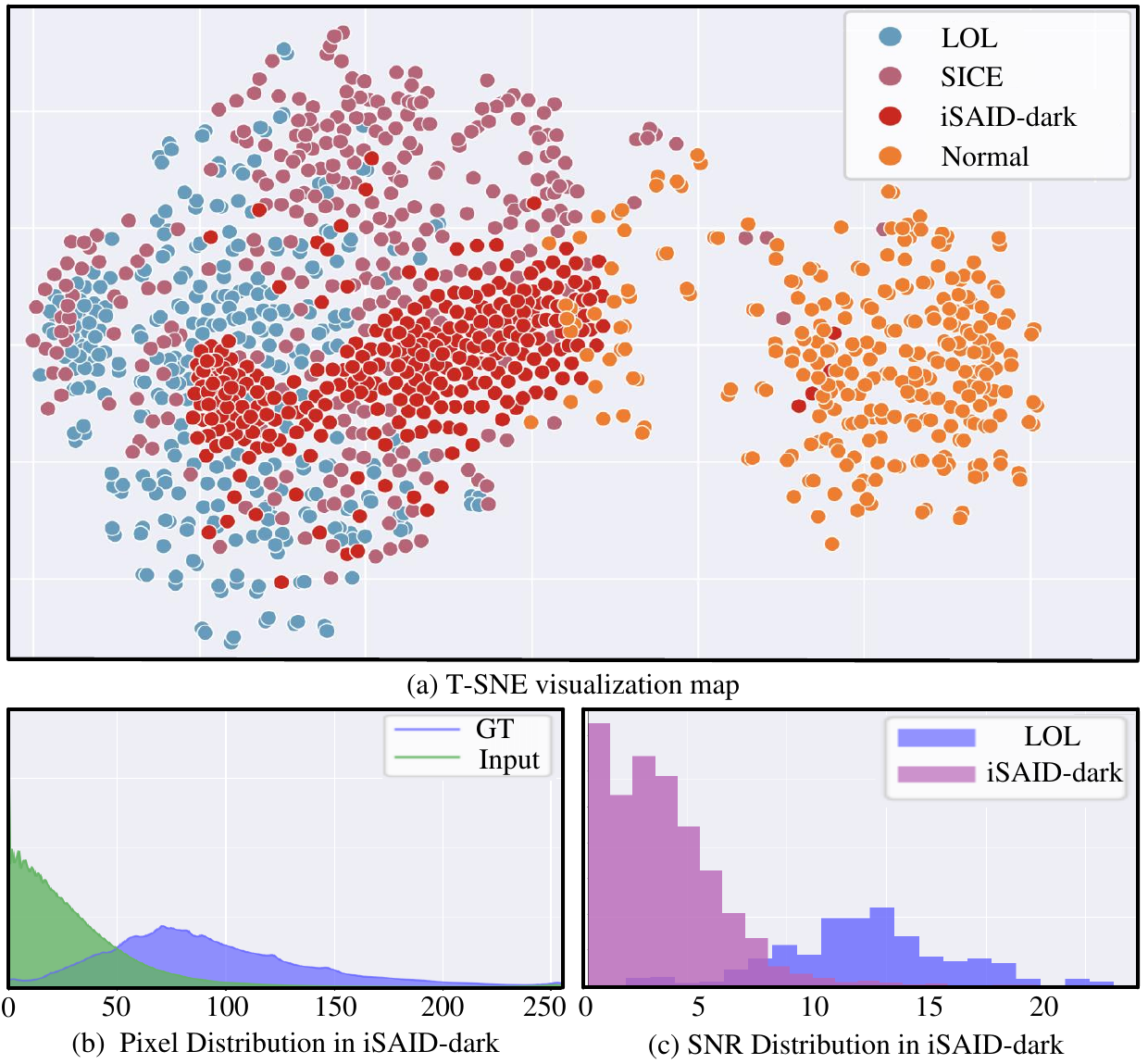}
	\caption{Distribution analysis of the iSAID-dark dataset. (a) T-SNE map indicates that the proposed dataset exhibits a broader data range. (b) demonstrates the brightness range of normal-light (blue) and low-light images (green). (c) In this figure, the range distribution of SNR values for both the LOL dataset and the proposed dataset is shown. It can be observed that the proposed dataset has a lower SNR range, indicating that its noise levels are more challenging.}
	\label{fig6}
\end{figure}

In summary, the amplitude illumination stage utilizes DDAB as the basic unit and employs an encoder-decoder structure to extract multi-scale feature information. Similarly, the phase refinement stage adopts DDPB as the basic unit. Skip connections are established between the corresponding encoder and decoder at each scale of the amplitude illumination stage. In contrast, the phase refinement stage employs IFAM to map features from the previous stage and the current encoder features more efficiently for fusing information between the two stages. Furthermore, to optimize the cross-stage network, a cross-stage feature fusion and supervised attention module (CSAM)\cite{zamir2021multi} is introduced at the end of the amplitude illumination stage.    

Next, we will provide a detailed explanation of the proposed IFAM module and Loss function.

	\subsection{Information Fusion Affine Module}
	
	Multi-stage networks\cite{fuoli2021fourier,li2021low,wang2023fourllie} enhance performance by decomposing complex problems into simpler sub-problems and solving them separately. In recent years, various works have achieved significant results using this approach. However, improving the ability of cross-stage information interaction remains a major challenge for multi-stage networks. As the depth of the network increases, shallow features tend to be lost gradually. Simply passing the output from the previous stage to the next stage may result in the loss of a significant amount of potential features. Another mainstream approach is to transfer information between stages of the same size, but this lacks interaction between features of different scales.

	Therefore, our designed IFAM first aggregates feature information from different stages and scales. Subsequently, it performs adaptive fusion on the aggregated features to enhance global contextual representation. Specifically, IFAM (as shown in Fig. \ref{fig4}) comprises two sub-modules:

		\subsubsection{Information Fusion Module}
		{
		      
		    IAM integrates all scale information from the first-stage decoder and second-stage encoder, and merges 	information from different scales. As shown in Fig. \ref{fig4} (a), the decoder features $ \left\{ A_{i} \right\}^{3}_{i=1} $ first undergo a 3 $\times$ 3 Conv, followed by combining different scale features through upsample and downsample operations. Then, 1 $\times$ 1 Conv is used to merge these features with different receptive fields, obtaining rich contextual information. The encoder features $\left\{P_{i} \right \}^{3}_{i = 1}$  undergo similar operations. Subsequently, the fused features are passed into IAM.
		}

		\subsubsection{Information Affine Module}
		{

		    \begin{table*}[htbp]
		        \centering
		        \caption{Comparison with state-of-the-art methods on the reference dataset, with the best results indicated 	in \textbf{bold} and the second-best results \uline{underlined}.}
		        \setlength\tabcolsep{4pt} 
		        
		        \begin{tabular}{|l|c||ccc|ccc|ccc|ccc|}
		            \hline 
		            \multicolumn{1}{|l|}{\multirow{2}{*}{\makecell{Methods}}} & 	\multicolumn{1}{c||}{\multirow{2}{*}{\makecell{Venue}}} & \multicolumn{3}{c|}{LOL} & \multicolumn{3}{c|}{iSAID-dark} & \multicolumn{3}{c|}{iSAID-dark (retrain)} & \multicolumn{3}{c|}{iSAID-dark (high-pixel)} \bigstrut[t]\\
		            \cline{3-14}
		            &   & PSNR↑ & SSIM↑ & LPIPS↓ & PSNR↑ & SSIM↑ & LPIPS↓ & PSNR↑ & SSIM↑ & LPIPS↓  & PSNR↑ & SSIM↑ & LPIPS↓ 	\bigstrut[t]\\
		            \hline
		            \hline
		            Zero-DCE++\cite{li2022learning} & TPAMI'22 & 12.17  & 0.511  & 0.258  & 12.07 & 0.040  & 0.818  & 8.46  	& 0.055 & 0.950 & 12.20 & 0.157 & 0.920 \bigstrut[t]\\
		            SCI\cite{ma2022toward}   & CVPR'22 & 14.78  & 0.525  & 0.238  & 12.14 & 0.194  & 0.901  & 13.08  & 0.205 	& 0.745 &13.73 &0.168 &0.854 \\
		            SNR\cite{xu2022snr}   & CVPR'22 & 21.22  & \textbf{0.834}  & 0.097  & 13.31 & 0.246  & 0.722  & 	\uline{23.17} & \uline{0.760}  & \uline{0.179} &\uline{22.06} &\uline{0.711} &\uline{0.238} \\
		            LLFormer\cite{wang2023ultra}  & AAAI'23 & \uline{23.25}  & 0.819  & 0.105  & \uline{16.66} & 	\uline{0.376}  & 0.658  & 23.11  & 0.725  & 0.221 &-&- &- \\
		            CUE\cite{zheng2023empowering}   & ICCV'23 & 21.68  & 0.769  & 0.151  & 15.70  & 0.343  & 0.717  & 21.47 	& 0.682 & 0.263 &19.16 &0.631 &0.349 \\
		            FourLLIE\cite{wang2023fourllie} & ACMM'23 & 20.02  & 0.818  & \uline{0.088} & 14.49 & 0.350  & 	\uline{0.579} & 19.78 & 0.643 & 0.309 &17.58 &0.503 &0.302 \\
		            LANet\cite{yang2023learning} & IJCV'23 & 21.74  & 0.816  & 0.101  & 11.88 & 0.089  & 0.804  & 18.72 & 	0.695 & 0.231 &21.95 &0.690 &0.247 \\
		            NeRCo\cite{yang2023implicit} & CVPR'23 & 22.94  & 0.783  & 0.102  & \textbf{16.73} & \textbf{0.513} & 	\textbf{0.381} & 18.79 & 0.604 & 0.326 &20.94 &0.568 &0.377 \\
		            \textbf{Ours} & -     & \textbf{23.46} & \textbf{0.834} & \textbf{0.087} & 15.51 & 0.287  & 0.714 & 	\textbf{25.30} & \textbf{0.784} & \textbf{0.151} &\textbf{22.47} &\textbf{0.721} &\textbf{0.192} \bigstrut[b]\\
		            \hline
		        \end{tabular}%
		        \label{tab1}%
		    \end{table*}%
	    
		    As shown in Fig. \ref{fig4} (b), taking the feature information from the i-th layer as an example, three 	features of the same size from different encoders and decoders are first fused through a 1$\times$1 Conv. Subsequently, they pass through separate 3 $\times$ 3 Conv to obtain spatial context information, and through a pooling layer and 1 $\times$ 1 Conv to obtain channel context information. Finally, the two streams of information are added element-wise to fully utilize their information.
		    \begin{equation}
		    \left\{
		    \begin{aligned}
		        &F_{fuse} = \Theta \left[\mho (A_{i}, P_{i}, U_{i}) \right],\\
		        &F^{'}_{fuse} = \Theta (F_{fuse}) + \text{Pool} \left\{\text{Reshape} \left[ \Theta(F_{fuse})\right]\right\},
		    \end{aligned}
		    \right.
		    \end{equation}
		    where $U_{i}$ represents the decoder features from the second stage.       
		
		     Subsequently, weight filters $F_{sv} \in  \mathbb{R}^{(k \times k \times c) \times h \times w} $ are generated based on 1 $\times$ 1 Conv, and reshaped into independent pixel kernels $F^{'}_{sv} = \left\{W_{i,j} \mid i \in [1,h], j \in [1,w] \right\}$, where $W_{i,j} \in \mathbb{R}^{k^{2}\times c} $. In this way, each position is adjusted to the optimum with adaptive weights. The filtering result can be represented as:
		    \begin{equation}
		        O_{i} = U_{i} \otimes F^{'}_{sv} + U_{i},
		    \end{equation}	
		    	where $\otimes$ indicates the element-wise multiplication. 
		    
		}

	Through the above process, IFAM generates adaptive weight filtering guided by multi-stage, multi-scale, and cross-domain interactive features, enabling the generated features to adapt more flexibly to image content.

	\subsection{Loss Function}
	    Due to our DFFN having two outputs $O_{A}$ and $O_{P}$, the loss functions $L_{A}$ and $L_{P}$ for the two stages are as follows:
     \begin{equation}
    \resizebox{\linewidth}{!}{%
    $\left\{
        \begin{aligned}
            &L_{A} = \left \| O_{A} - \mathcal{F}^{-1}(\mathcal{A}(I_{gt}),\mathcal{P}(I_{low}) \right \|_{1} + \alpha \left \|\mathcal{A}(O_{A}) - \mathcal{A}(I_{gt}) \right \|_{1},\\
            &L_{P} = \left \| O_{P} - I_{gt} \right \|_{1} + \beta \left \|\mathcal{F}(O_{P}) - \mathcal{F}(I_{gt}) \right \|_{1}  + \gamma  \left \|\mathcal{P}(O_{P}) - \mathcal{P}(I_{gt}) \right \|_{1},
        \end{aligned}
    \right.$%
    }
    \end{equation}
    where $\left \| \cdot \right \|_{1}$ presents the Mean Absolute Error (MAE), $\alpha$, $\beta$ and $\gamma$ are the weight factor, and we empirically set them to 0.05. Therefore, the overall loss is :
    \begin{equation}
        L = L_{A} + L_{P}.
    \end{equation}

\section{dataset}

	\begin{figure*}[!t]
		\centering
		\includegraphics[width=2\columnwidth]{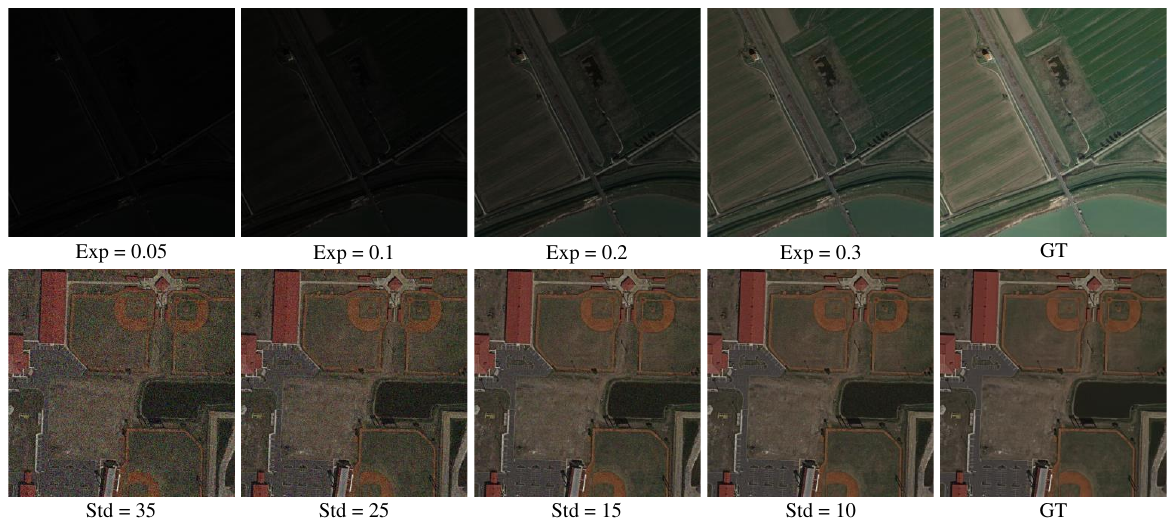}
		\caption{The proposed iSAID-dark dataset contains low-light images with varying exposure levels and noise levels.}
		\label{fig7}
	\end{figure*}

	Due to the particularity of remote sensing images, collecting paired normal/low-light image sets is almost an impossible task. In this work, inspired by the significant achievements of previous synthetic datasets in various low-quality image enhancement tasks, we introduced a synthesis mechanism to generate the corresponding low-light remote sensing dataset. Additionally, to validate the reliability of the proposed DFFN and iSAID-dark in real low-light environments, we collected real nighttime remote sensing images by capturing different scenes with drones at night, forming a test dataset named dark remote sensing (darkrs), which comprises 86 test images. The two proposed datasets are shown in Fig. \ref{fig5}.

\subsection{Data Collection}

   The iSAID\cite{waqas2019isaid} dataset has collected a large number of high-resolution aerial images from various scenes including rural and urban areas. Therefore, we selected 751 images from the iSAID test set as our benchmark data. Considering the significant differences in size and shape of remote sensing images, and the difficulty of training high-resolution images in mainstream networks, and in order to increase the diversity of scenes included in the dataset, we performed multiple random crops on these 751 images. The cropped images are resized to 500x500. Subsequently, we divided them into 3755 training image pairs and 66 validation image pairs. To further validate the recovery performance at high resolutions, we constructed a corresponding high-resolution dataset called iSAID-dark (high-pixel), which includes 72 test images, all with a resolution of 1080p.

	
   \begin{table*}[htbp]
   	\centering
   	\caption{Comparison with state-of-the-art LLIE methods on the no-reference datasets, along with the number of parameters, GFLOPS, and FPS on images of 1080p resolution. The best results are indicated in \textbf{bold}, and the second-best results are \uline{underlined}.}
   	\setlength\tabcolsep{9pt}
   	\begin{tabular}{|l|c||c|c|c|c|c|c|c|c|c|}
   		\hline
   		Methods & Venue & \multicolumn{1}{c|}{Exdark} & \multicolumn{1}{c|}{DICM} & \multicolumn{1}{c|}{NPE} & \multicolumn{1}{c|}{LIME} & \multicolumn{1}{c|}{darks} & \multicolumn{1}{c|}{Ave.} &FPS&FLOPs(G)& Params (M) \bigstrut\\
   		\hline
   		\hline
   		Zero-DCE++\cite{li2022learning} & TPAMI'22 & 3.992 & 3.924 & \uline{3.489} & 4.188 & 3.760 & \uline{3.870} & \uline{281} &\textbf{0.21} & \uline{0.0106} \bigstrut[t]\\
   		SCI\cite{ma2022toward}   & CVPR'22 & 3.968 & \uline{3.657} & 3.912 & 4.165 & 5.450 & 4.230 & \textbf{293} & \uline{1.1} &  \textbf{0.0004} \\
   		SNR\cite{xu2022snr}   & CVPR'22 & 4.125 & 4.458 & 5.503 & 5.844 & \uline{3.456} & 4.677 &3 & 880 & 40.0842 \\
   		LLFormer\cite{wang2023ultra}  & AAAI'23 & 3.960 & 3.890 & 3.629 & 4.175 & 4.502 & 4.031 & -& - &  24.5183 \\
   		CUE\cite{zheng2023empowering}   & ICCV'23 & 3.989 & 3.722 & 3.977 & 4.068 & 3.708 & 3.892 & 3 & 198 &  0.2407 \\
   		FourLLIE\cite{wang2023fourllie} & ACMM'23 & \uline{3.941} & 3.685 & 3.550  & 4.056 & 4.119 &  \uline{3.870} & 6&515 & 1.4855 \\
   		LANet\cite{yang2023learning} & IJCV'23 & 4.322 & 3.992 & 3.828 & 4.493 & 3.478 & 4.022 & 15& 317 &  0.0422 \\
   		NeRCo\cite{yang2023implicit} & CVPR'23 & 4.092 & 4.052 & 4.048 & \uline{4.007} & 4.140  & 4.067 & 3& 1341 &  7.8414 \\
   		Ours  & -     & \textbf{3.673} & \textbf{3.540} & \textbf{3.347} & \textbf{4.003} & \textbf{3.258}  & \textbf{3.564} & 4 &501 &  2.5765 \bigstrut[b]\\
   		\hline
   	\end{tabular}%
   	\label{tab2}%
   \end{table*}%

    \subsection{Data Generation}
        \subsubsection{Low-light Simulation}
            Inspired by Zero-DCE++\cite{li2022learning}, which treats enhancing illumination as an estimation task of specific curves for images and achieves excellent results, we adopted an inverse transformation of Zero-DCE++ to simulate low-light images by adjusting the reverse curve. This allows us to generate low-light images with adjustable darkness levels. Specifically, we set a random darkness range to replace the fixed exposure value of brightness loss in Zero-DCE++. By using different exposure levels, the trained reverse Zero-DCE++ can generate low-light images with varying darkness levels.
        \subsubsection{Noise Simulation}
        To simulate more realistic low-light images, we also added Gaussian noise of varying degrees to the generated images, all of which followed a normal distribution. Additionally, for images with higher levels of darkness, the added noise was more intense.
        
		\begin{figure*}[!t]
			\centering
			\includegraphics[width=0.98\textwidth]{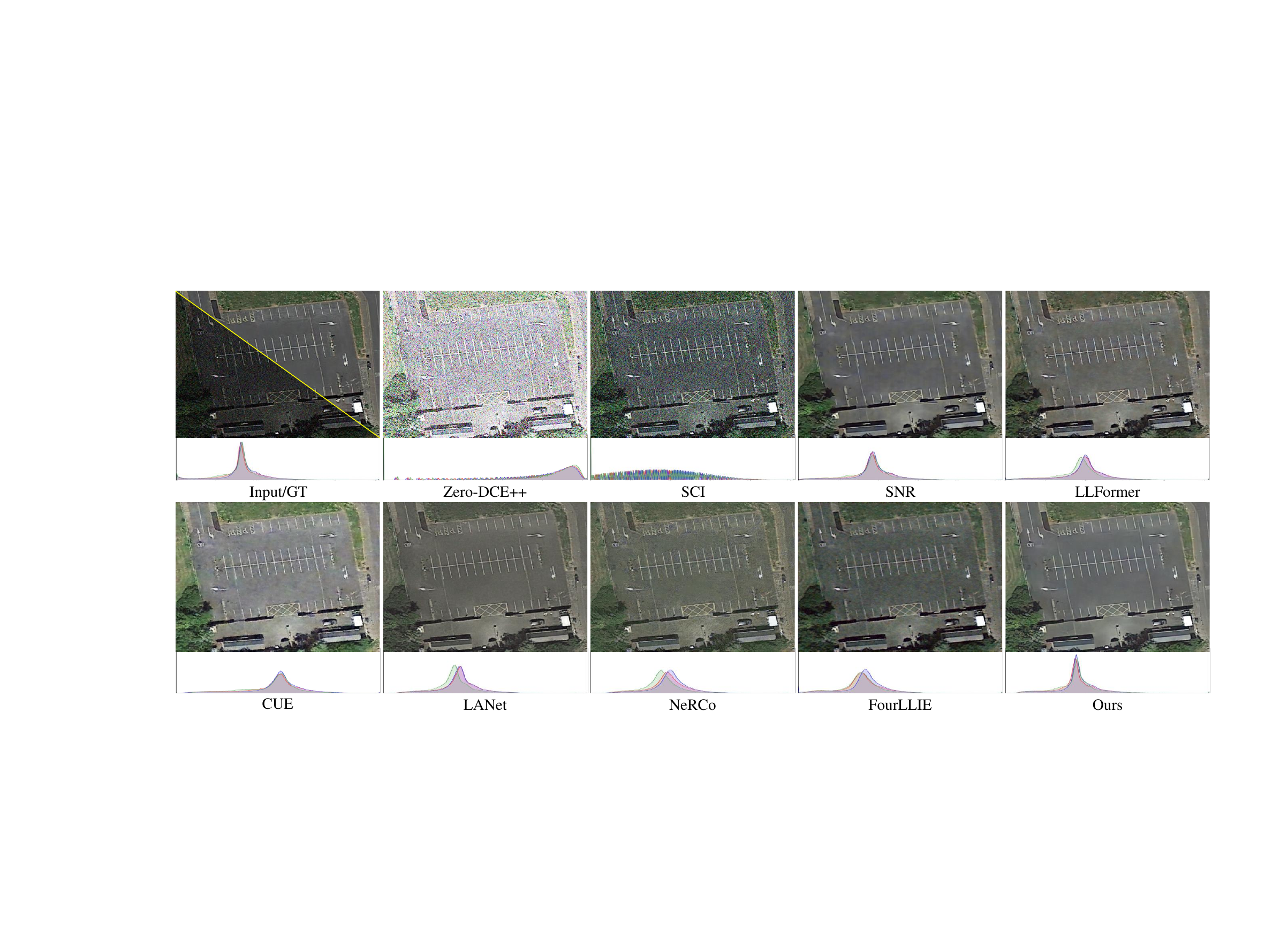}
			\caption{The visualization results on the iSAID-dark dataset. We present the histogram of color distribution for the images. The histograms placed in Input/GT represent the color distribution of the GT. It can be observed that our method's histogram is closer to the GT histogram.}
			\label{fig8}
		\end{figure*}  
	
    \subsection{Data Distribution}
        Fig. \ref{fig6} (a) displays the T-SNE visualization results of our proposed iSAID-dark dataset compared to other datasets, demonstrating that iSAID-dark encompasses common features of LOL and SICE, indicating a broader data range in our dataset. Fig. \ref{fig6} (b) illustrates the brightness distribution of our proposed dataset.  Meanwhile, we plotted the SNR distribution of the proposed dataset and the LOL dataset, as shown in Fig. \ref{fig6} (c), to demonstrate the noise levels in the images. The SNR distribution indicates that our dataset possesses a wide and challenging SNR range.

    \subsection{Discussion}

        Our dataset simulates real low-light environments, encompassing the majority of common remote sensing scenes. Additionally, our randomly set dark intensity range and noise cover most common challenging low-light scenarios. When given different exposure strengths, the reverse Zero-DCE++ can generate low-light images with different brightness levels. Fig. \ref{fig7} illustrates images with different levels of low-light and noise.

    \section{Experiments}

   \begin{figure*}[!t]
   	\centering
   	\includegraphics[width=0.98\textwidth]{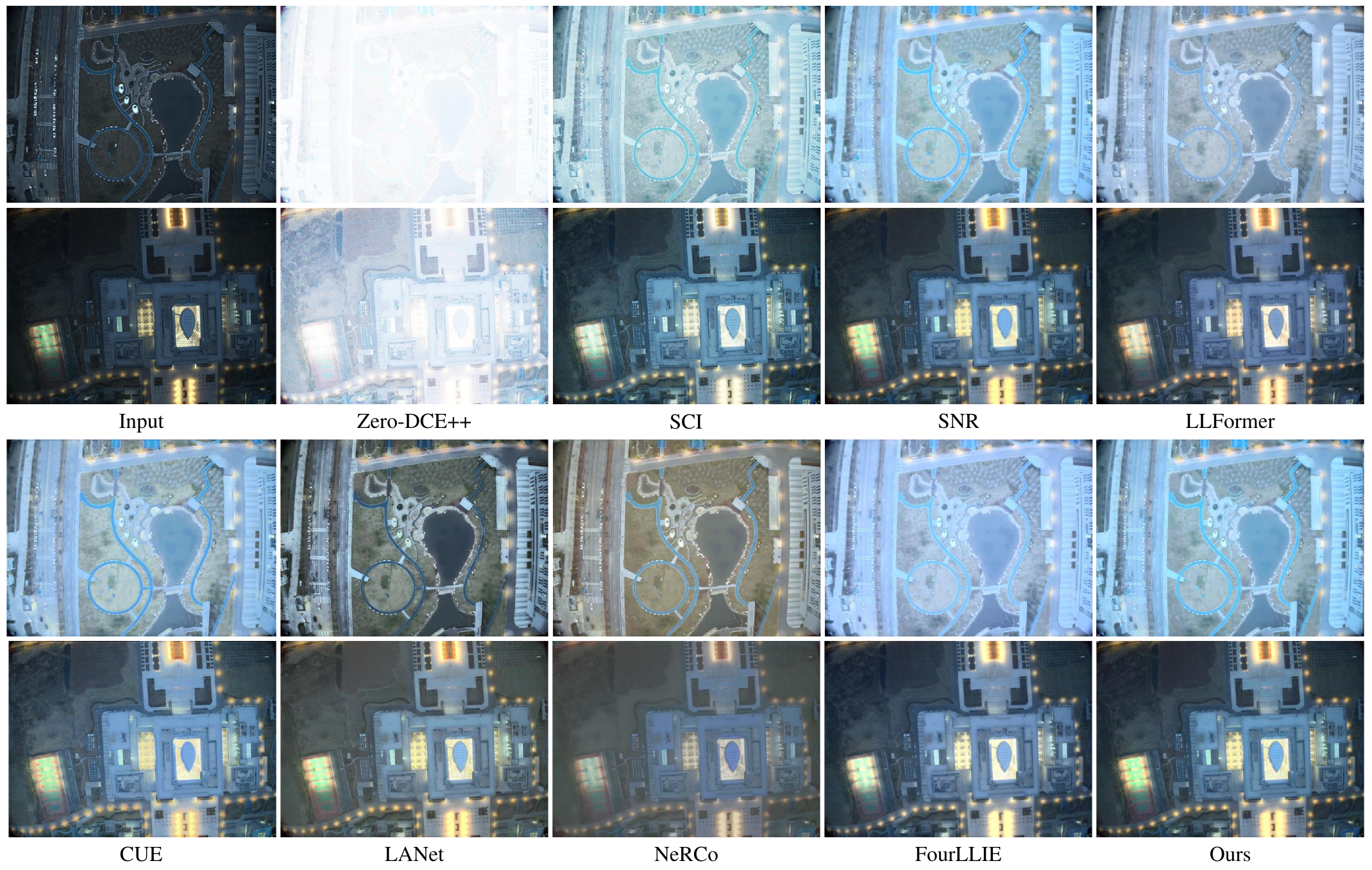}
   	\caption{The visualization results on the darkrs dataset.}
   	\label{fig9}
   \end{figure*}

    In this section, we first describe the detailed setup of the experiments, including the datasets used, evaluation metrics, comparison methods, and implementation details. Subsequently, we quantitatively and qualitatively evaluate the proposed DFFN against state-of-the-art methods on the proposed low-light remote sensing dataset and real-world low-light dataset. Additionally, we provide extensive ablation studies to validate the effectiveness of the proposed DFFN architecture and its components.
    
 	\subsection{Experiment Details}

 		\subsubsection{Datasets and Evaluation Metrics}
 			We first validate the performance of the proposed method on the iSAID-dark and dark-rs datasets. Additionally, to assess the generalization ability of our model to low-light images in other scenarios, we further evaluate the performance of the proposed method on five widely used datasets, including a reference-based dataset (LOL\cite{wei2018deep} ) and three no-reference datasets (DICM, Exdark, and NPE).
 			
 			We evaluate the visual quality of reference-based datasets using PSNR\cite{huynh2008scope}, SSIM\cite{wang2004image}, and LPIPS\cite{zhang2018unreasonable} metrics. For no-reference images, we assess their naturalness using the NIQE\cite{mittal2012making} metric. Higher PSNR and SSIM scores indicate better quality, while lower NIQE and LPIPS scores are desirable.

 		\begin{figure*}[!t]
 			\centering
 			\includegraphics[width=0.98\textwidth]{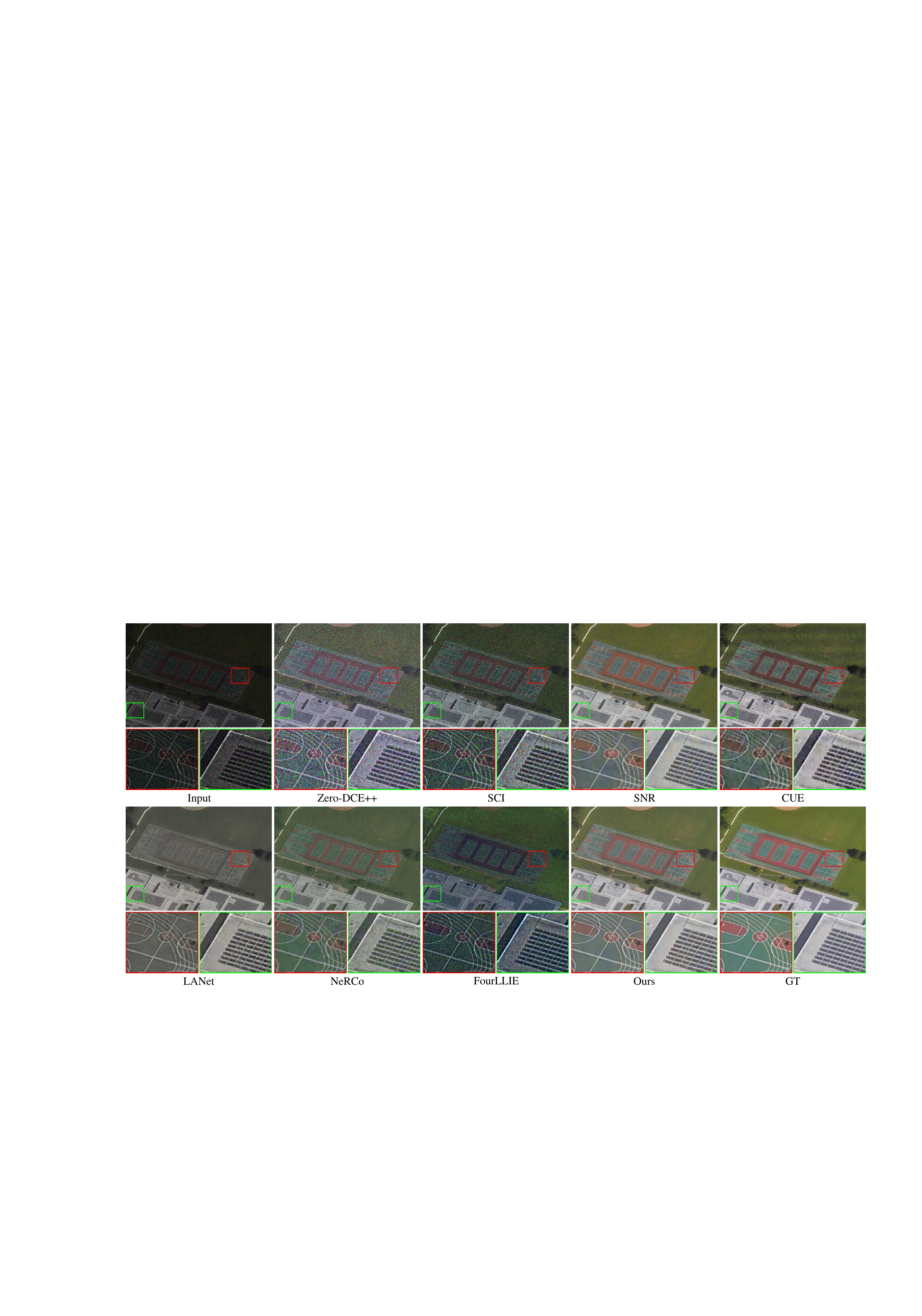}
 			\caption{ The visual results on the 1080p image. The selected area is magnified to show details.}
 			\label{fighigh}
 		\end{figure*}

 		\begin{figure*}[!h]
 			\centering
 			\includegraphics[width=0.98\textwidth]{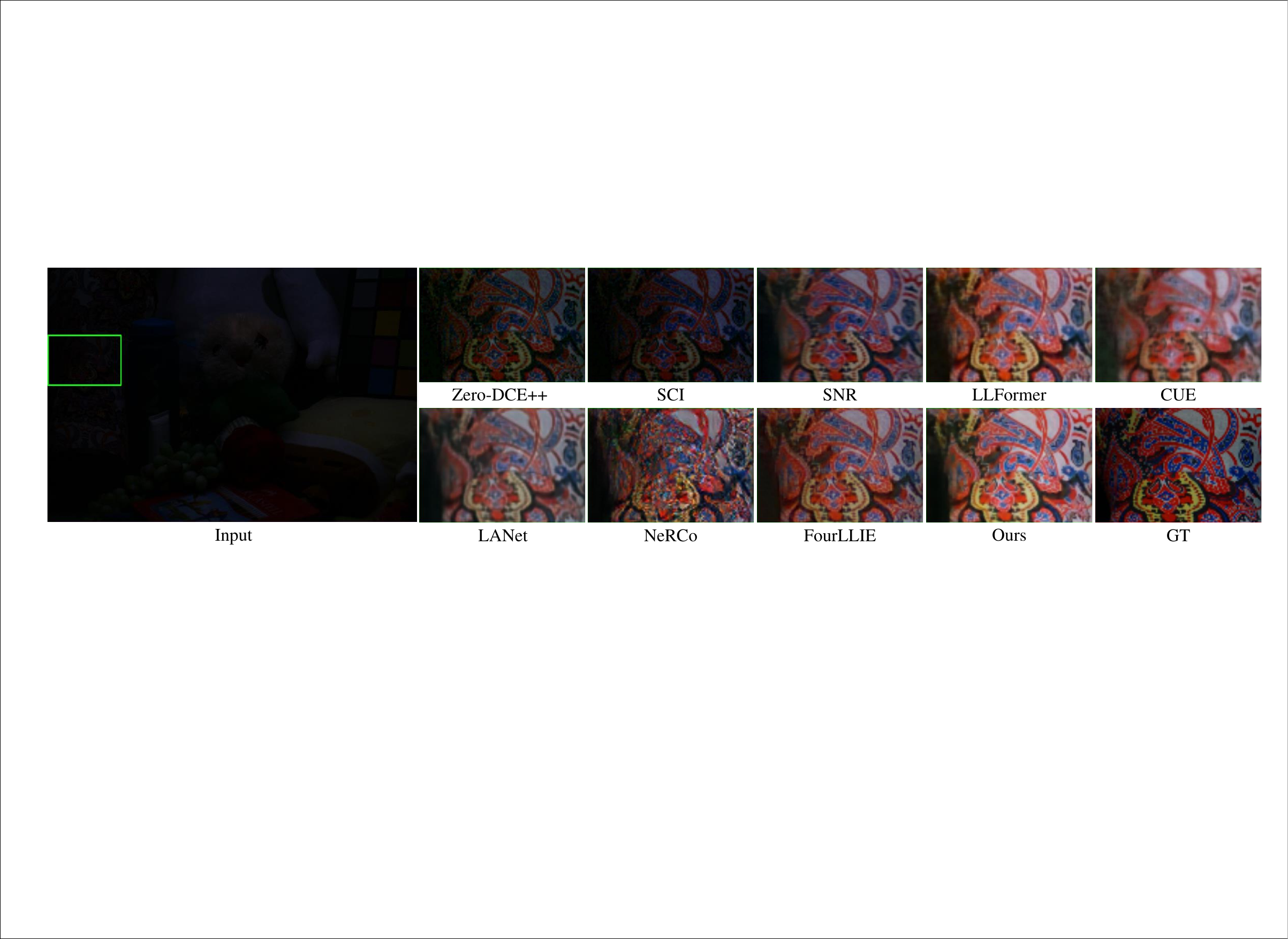}
 			\caption{The visualization results on the LOL dataset. The selected area is magnified to show details.}
 			\label{fig10}
 		\end{figure*}	
   
 		\begin{figure*}[!h]
 			\centering
 			\includegraphics[width=0.98\textwidth]{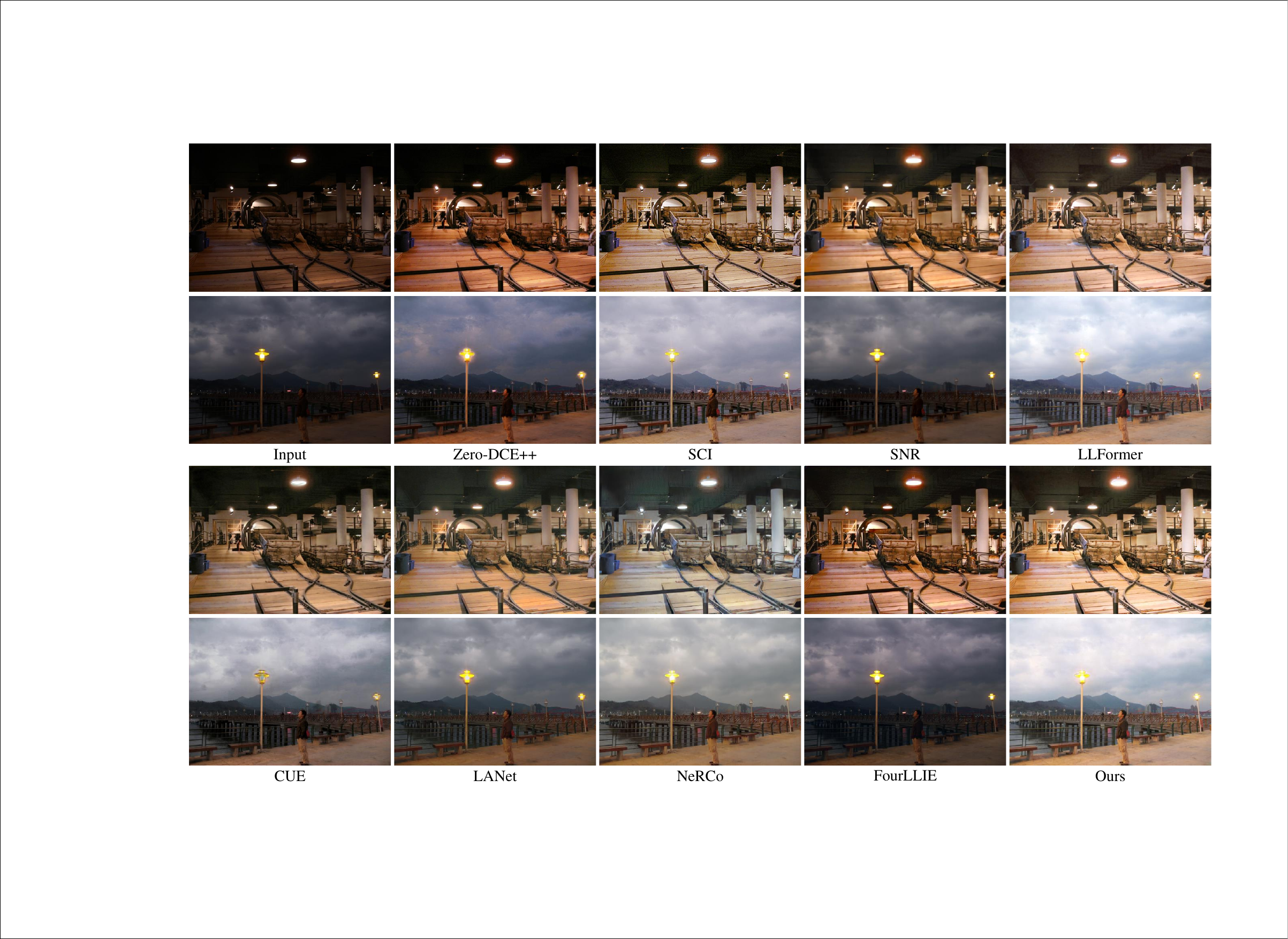}
 			\caption{The visualization results on the DICM dataset (top) and the NPE dataset (bottom).}
 			\label{fig11}
 		\end{figure*}

		 \begin{figure*}[!t]
			\centering
			\includegraphics[width=0.98\textwidth]{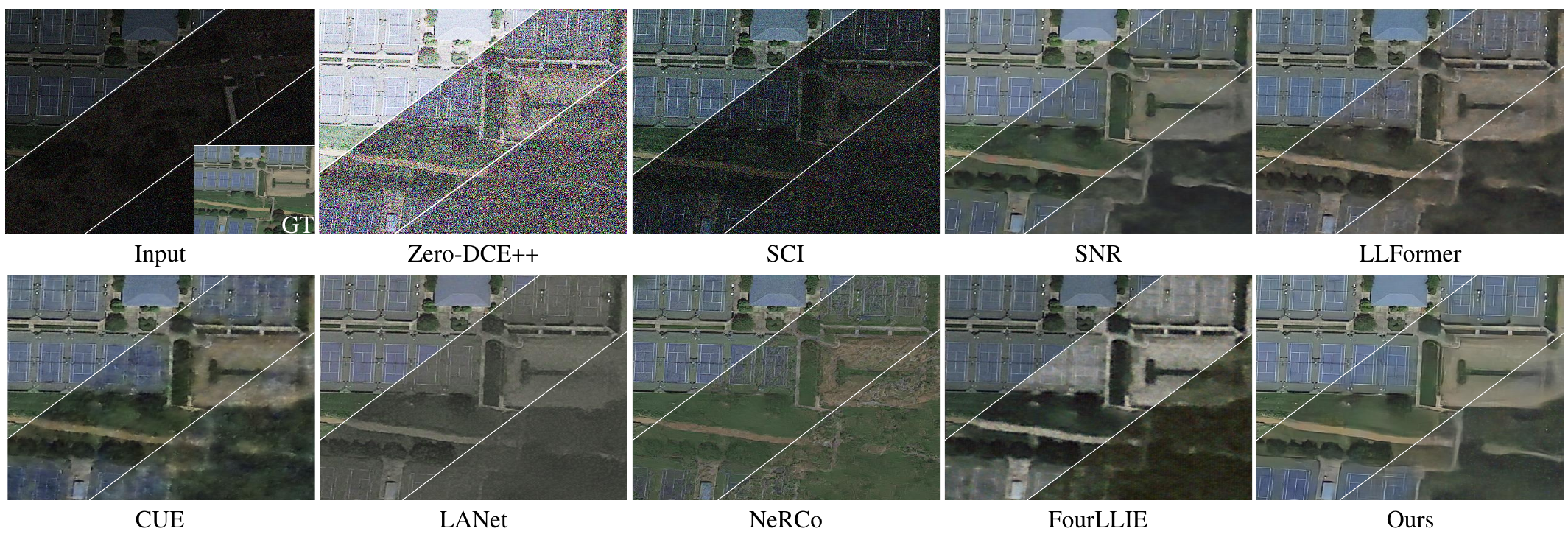}
			\caption{Qualitative enhancement results of three low-light images of the same scene. From left to right, the brightness of the images gradually decreases, and the noise concentration gradually increases.}
			\label{fimutilevel}
		\end{figure*}
 		\subsubsection{Compared Methods}
 			To evaluate the performance of the proposed method, we compare DFFN with current state-of-the-art methods, including a curve-based method (Zero-DCE++\cite{li2022learning}), two unsupervised learning-based methods (SCI\cite{ma2022toward}, NeRCo\cite{yang2023implicit}), and five supervised learning-based methods (LLFormer\cite{wang2023ultra}, SNR\cite{xu2022snr}, CUE\cite{zheng2023empowering}, FourLLIE\cite{wang2023fourllie}, and LANet\cite{yang2023learning}). In the experiments, we retrain and test these methods using the weights and publicly available code provided by these approaches (for the proposed iSAID-dark dataset) for fair comparison.
		\subsubsection{Implementation Details}
			
			Our model is trained on the LOL dataset and the iSAID-dark dataset separately. To facilitate better convergence of the model, we set the initial learning rate to 0.0002, and then linearly decrease it by half every 100 epochs, with a total of 500 epochs for training. We use the ADAM optimizer for model optimization. All experiments are conducted on a Windows computer equipped with an NVIDIA RTX 3090 (24GB RAM) for training and testing. Additionally, we set the batch size to 12, randomly crop each image into 256$\times$256 patches, and augment these patches using data augmentation techniques such as horizontal and vertical flips. For image features, we set C to 20, and subsequently, the feature levels are transformed to 40, 80, and 160.
			
	\subsection{Comparison With Stat-of-the-Art Methods}
		\subsubsection{Quantitative Results}

			We first quantitatively compare DFFN with state-of-the-art methods. Table I presents the PSNR, SSIM, and LPIPS results on reference-based datasets. 
			From Table I, it can be seen that our method achieved the best results on both reference datasets. It is worth noting that the results shown for iSAID-dark are obtained by testing with weights trained on LOL, indicating that models trained on existing datasets may not be suitable for low-light remote sensing image enhancement. 
			
			Table II presents the NIQE results on the no-reference dataset. It can be observed from Table II that our method achieves the best  on all datasets, which fully demonstrates the superiority of DFFN.	
			We also present the parameter count of the models in Table . The results indicate that although our method cannot compete with no-reference methods in terms of complexity, the performance of no-reference methods on low-light remote sensing datasets is extremely unsatisfactory (for example, SSIM of Zero-DCE++ on iSAID-dark is only 0.055). In contrast, our method achieves a remarkable balance between model complexity and performance.

		\begin{figure*}[!t]
				\centering
				\includegraphics[width=0.98\textwidth]{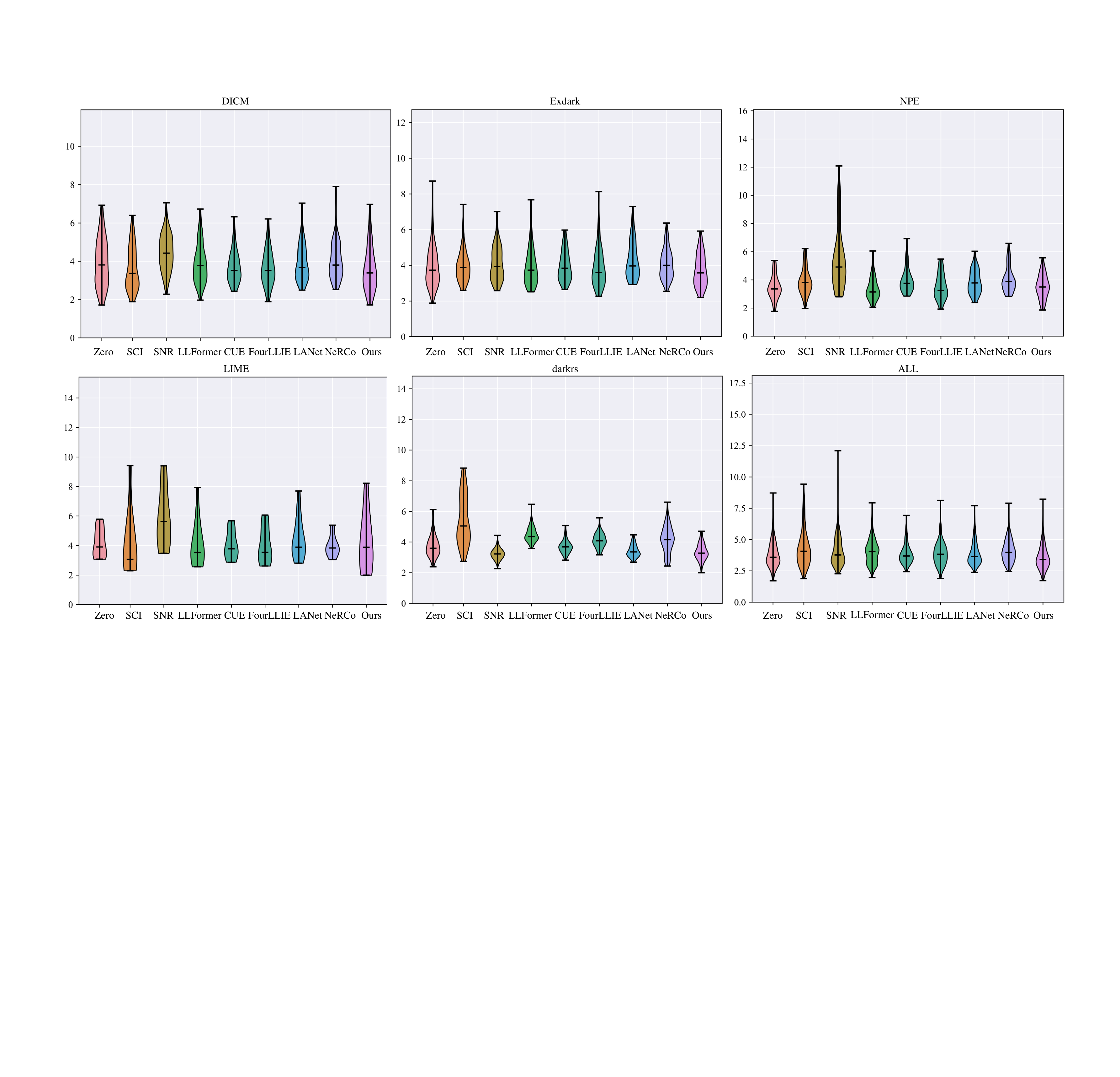}
				\caption{On no-reference image datasets, the numerical distribution of NIQE test results varies across different enhancement methods. The evaluation value of the proposed method is relatively concentrated, with low scores across multiple images, and better performance in low-light enhancement scenarios.}
				\label{fig12}
		\end{figure*}

		\subsubsection{Qualitative Results}
		
			Fig. \ref{fig8} illustrates the enhancement results of our method compared to other state-of-the-art methods on the iSAID-dark dataset. Additionally, histograms of color distributions for these images are plotted, where the histograms shown under Input/GT represent the GT histograms. Due to the significant Gaussian noise added to the iSAID-dark dataset, no-reference methods such as Zero-DCE++ and SCI, which were not designed to address denoising issues, still exhibit substantial noise in the recovered images. Consequently, these two methods show low SSIM and PSNR values as indicated in Table I. Furthermore, the restoration results of other methods either suffer from excessive color distortion or exhibit blurred details due to noise. In contrast, our method achieves the best visual results, while the histograms of color distributions are closer to the GT.
			
			Fig. \ref{fig9} depicts remote sensing images captured by drones in real nighttime environments. The dataset contains numerous areas with uneven exposure, providing a better balance for evaluating the model's ability to enhance low-light remote sensing images in real-world scenarios. It can be observed that Zero-DCE++ exhibits overexposure, while LANet, and NeRCo show limited brightening effects, resulting in overall low contrast in the images. SCI, LLFormer and FourLLIE demonstrate good enhancement in low-light areas, but some regions suffer from color distortion, and FourLLIE blurs image details. Our method, effectively enhances image brightness while better preserving image color.
			
			As shown in Fig. \ref{fighigh}, for high-resolution images, Zero-DCE++, SCI and FourLLIE struggle to
			restore the image brightness and fail to address noise issues. CUE, LANet, SNR, and NeRCo
			exhibit unexplained artifacts or neglect color information. LLFormer, due to its large computational load, cannot process images of this resolution on our equipment. In contrast, our method maintains outstanding performance on high-resolution image.
			
			Fig. \ref{fig10} displays the enhancement results on the LOL dataset. We have selected certain regions within the images for zooming in to present the image details more clearly. From the image, it can be seen that Zero-DCE++, SCI, and SNR only achieved limited enhancement. The results of CUE and LANet are too blurred in details. NeRCo exhibited some unexplained shadows. Although FourLLIE and LLFormer also achieved decent enhancement effects, compared to them, our method shows more prior colors and clearer details.

			\begin{table}[htbp]
				\centering
				\caption{Ablation study of the two-stage architecture design includes experiments where 'Frequency' indicates whether frequency domain is used, 'Stage2' denotes whether the second stage is utilized, and 'Swap' indicates whether the phase information of the output $O_{1}$ from the first stage is replaced.}
				\begin{tabular}{|c||c|c|c|c|c|c|}
					\hline
					Model & Frequency & Stage2 & Swap  & PSNR↑ & SSIM↑ & LPIPS↓ \bigstrut\\
					\hline
					\hline
					$M_{a}$  & \ding{55}     & \ding{55}     & \ding{55}     & 21.58 & 0.734 & 0.198 \bigstrut[t]\\
					$M_{b}$  & \checkmark     & \ding{55}     & \ding{55}     & 22.59 & 0.750  & 0.178 \\
					$M_{c}$  & \checkmark     & \checkmark     & \ding{55}     & 24.52 & 0.771 & 0.167 \\
					$M_{d}$  & \checkmark     & \checkmark     & \checkmark & \textbf{24.93} & \textbf{0.772} & \textbf{0.166} \bigstrut[b]\\
					\hline
				\end{tabular}%
				\label{tab3}%
			\end{table}%
			

		Fig. \ref{fig11} illustrates the enhancement results on the DICM dataset and the NPE dataset.  Zero-DCE++, SNR, and FourLLIE produce overly dark enhancement results, with little difference compared to the degraded images, especially in the images of the second row. SCI, CUE, LANet, and NeRCo introduce some color deviations and inadequate brightness restoration in the enhanced images. In comparison, our method effectively suppresses noise while restoring image brightness, presenting the most natural images.
			
		Meanwhile, we tested the recovery ability of the method in different lighting and noise environments, as shown in Fig. \ref{fimutilevel}. Zero-DCE++ and SCI exhibit significantly different recovery capabilities under varying illumination conditions, and their noise repair ability is poor. LANet and FourLLIE struggle to capture color information in extremely dark environments. Due to excessive noise interference, the detail recovery capabilities of CUE and NeRCo are greatly affected. Although LLFormer and SNR, based on the Transformer architecture, demonstrate relatively excellent hierarchical correction capabilities, their performance is still unsatisfactory under excessively dark and noisy conditions. In contrast, our method achieves excellent recovery results in all three levels of adverse environments.
			
		\begin{figure}[!h]
			\centering
			\includegraphics[width=1.0\columnwidth]{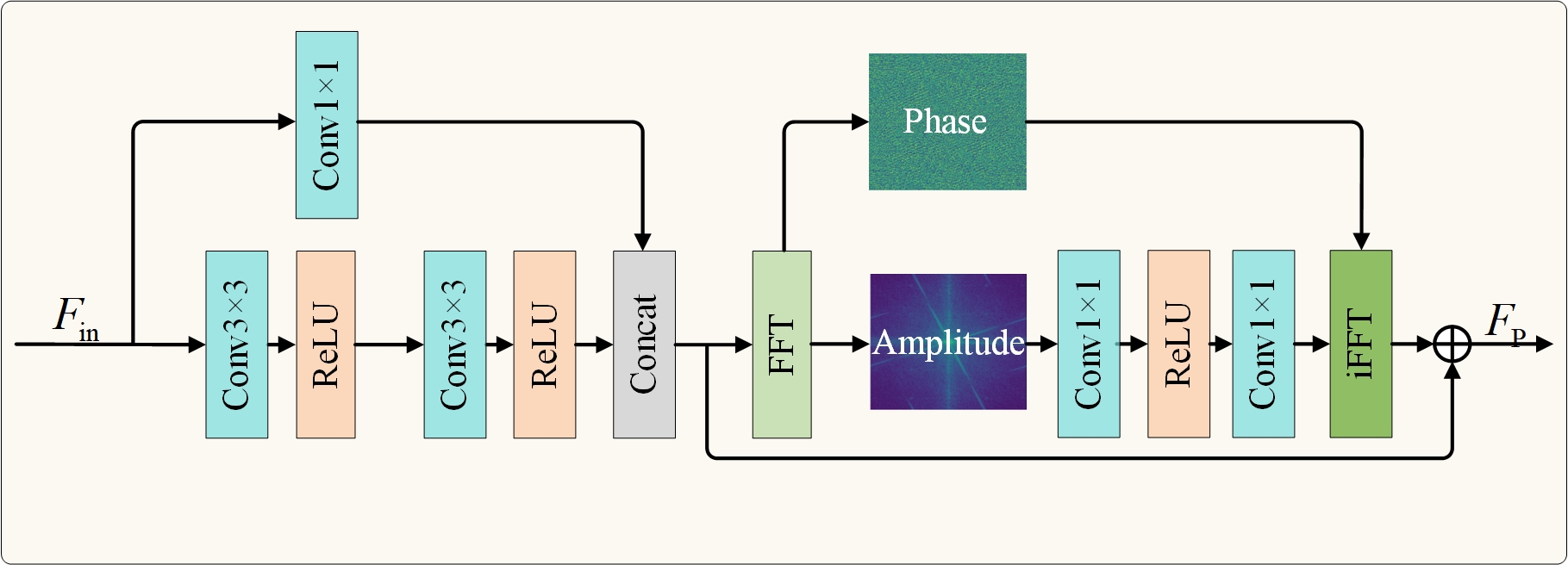}
			\caption{The modified DDAB first learns spatial domain information in a serial manner, followed by learning frequency domain information.}
			\label{fig13}
		\end{figure}
			
            \begin{table}[htbp]
				\centering
				\caption{Ablation study about the combination of spatial-frequency domain features.}
				\setlength\tabcolsep{7pt} 
				\begin{tabular}{|c||c|c|c|c|c|c|}
					\hline
				\multicolumn{1}{|c||}{Model} & \multicolumn{1}{c|}{$M_{1}$} & \multicolumn{1}{c|}{$M_{2}$} & \multicolumn{1}{c|}{$M_{3}$} & \multicolumn{1}{c|}{$M_{4}$} &\multicolumn{1}{c|}{$M_{5}$} &\multicolumn{1}{c|}{Ours} \bigstrut[t]\\
					\hline
					\hline
					PSNR↑  & 21.55 & 24.88 & 24.88 & 24.93 &22.42& \textbf{25.38}\bigstrut[t]\\
					SSIM↑  & 0.655 & 0.771 & 0.772 & 0.772 &0.701& \textbf{0.773}\bigstrut[t]\\
					LPIPS↓ & 0.289 & \textbf{0.162} & 0.171 & 0.166 &0.237& 0.165\bigstrut[t]\\
					\hline
				\end{tabular}%
				\label{tab4}%
			\end{table}%
		Fig. \ref{fig12} displays the NIQE distributions of the comparison methods across five test datasets. It is evident that DFFN exhibits a more concentrated data distribution, which strongly indicates the superior generalization capability of our model across various scenarios. Additionally, when all test set data is combined and plotted together (denoted as ALL), DFFN's data appears more centralized. This further demonstrates the exceptional stability of our model.

        \begin{table}[htbp]
				\centering
				\caption{The ablation study for networks with different depths. C = 10 means the initial feature channels are 10, which then expand sequentially to 2C, 4C, and 8C within the network.}
                \setlength\tabcolsep{3pt} 
				\begin{tabular}{|c||c|c|c|c|c|c|c|}
					\hline
					Depth & PSNR↑ & SSIM↑ & LPIPS↓ &FPS&FLOPs(G)&Params(M)\bigstrut[t]\\
					\hline
					\hline
					C = 10   & 21.58 & 0.734 & 0.222 &\textbf{7} &\textbf{128} &\textbf{0.6412}\bigstrut[t]\\
					C = 20  &  \textbf{25.30} & \textbf{0.784}  &\textbf{0.151} &4&501&2.5765 \\
					C = 30  &  24.90 & 0.735 & 0.176 &2&1121&5.7883\\
					\hline
				\end{tabular}%
				\label{tab6}%
		\end{table}%
		
		\begin{table*}[htbp]
			\centering
			\caption{The ablation study for using transformer blocks, Single Transformer Block (STB) indicates that each block uses only one self-attention block and one FFN, while Multiple Transformer Block (MTB) signifies that multiple self-attention blocks and FFNs are stacked within each transformer block.}
			\setlength\tabcolsep{10pt} 
			
				\begin{tabular}{|c||c|c|c|c|c|c|c|c|}
					\hline
					Model & Attention  & Head  & PSNR↑ & SSIM↑ & LPIPS↓ &FPS& FLOPs(G) & Params (M)\bigstrut[t]\\
					\hline
					\hline
					STB    & [1,1,1,1] & [2,2,2,2]   &23.53 & 0.687 & 0.260 &1 &1081& 6.3139\bigstrut[t]\\
					MTB    & [2,4,6,8] & [2,2,4,4]  & 23.02 & 0.726 & 0.203  &-&-& 26.9895 \bigstrut[t]\\
					Ours    &   -   &   -   & \textbf{25.38} & \textbf{0.788} & \textbf{0.146} &\textbf{4} &\textbf{501}&\textbf{2.5765} \bigstrut[t]\\
					\hline
			\end{tabular}%
			\label{tab5}%
		\end{table*}%
	
		\subsection{Ablation Study}
		\label{Ablation}
		In this subsection, we conducted extensive ablation experiments to validate the effectiveness of the proposed modules, the two-stage network, and the fusion of spatial and frequency domains. Subsequently, we evaluated the performance of networks with different depths and the use of multi-level loss. Finally, we replaced the frequency domain block with the state-of-the-art transformer block to validate the effectiveness of the proposed model.
		
		\subsubsection{Contribution of Two-stage Architecture}	
		In this subsection, we validate the effectiveness of the proposed two-stage network architecture. All experiments are conducted based on the foundational model without employing the IFAM modules. $M_{a}$ represents the use of spatial information only in a single stage, $M_{b}$ denotes the simultaneous utilization of spatial and frequency domain information in the single stage, and $M_{c}$  employs the two-stage architecture, but with $O_{1}$ as the input for the second stage, $M_{d}$ is identical to our training architecture
	
		As shown in Table \ref{tab3}, the performance of $M_{a}$ is the poorest, indicating that solely recovering spatial information in a single stage may yield unsatisfactory results. Compared to $M_{b}$, $M_{c}$ and $M_{d}$ exhibit significantly improved performance, implying that decoupling and separately learning degraded information are crucial for information recovery. Moreover, $M_{d}$ achieves a gain of 0.21dB in PSNR over $M_{c}$, indicating that preserving phase consistency is more critical for learning in the second stage.
		
		\subsubsection{Contribution of Dual-Domain Fusion}
		In this subsection, we discuss the effectiveness of integrating spatial and frequency domain information and the effectiveness of different integration methods, while also exploring the effectiveness of the proposed IFAM and DDAB/DDPB. $M_{1}$ denotes the use of spatial information only in a two-level network. $M_{2}$ and $M_{3}$ adopt a sequential method to learn spatial and frequency domain information, respectively. Specifically, $M_{2}$ first learns spatial information (as shown in Fig. \ref{fig13}), while $M_{3}$ first learns frequency domain information. $M_{4}$ has the same training architecture as ours, but none of the four configurations use IFAM. $M_{5}$ indicates the absence of the proposed DDAB/DDPB.
	
		As shown in Table \ref{tab4}, the model $M_{1}$	utilizing only spatial information performs poorly, indicating the effectiveness of integrating spatial-frequency dual-domain information. Compared to $M_{2}$ and $M_{3}$ ,$M_{4}$, which learns in parallel, achieves the best performance in most metrics. This suggests that compared to the parallel approach, serial learning is more prone to losing information from one domain while learning information from another. By employing parallel computation, this information loss is effectively mitigated. When DDAB/DDPB is not employed, the model's performance declines significantly, highlighting the superiority of the proposed DDAB/DDPB.
		
		Moreover, when IFAM is utilized, there is a significant improvement in PSNR, indicating that IFAM successfully aggregates features from different sources and enhances the global context representation of the network.
		
		\subsubsection{Contribution of different depths}
\begin{figure}[!h]
		\centering
		\includegraphics[width=1.0\columnwidth]{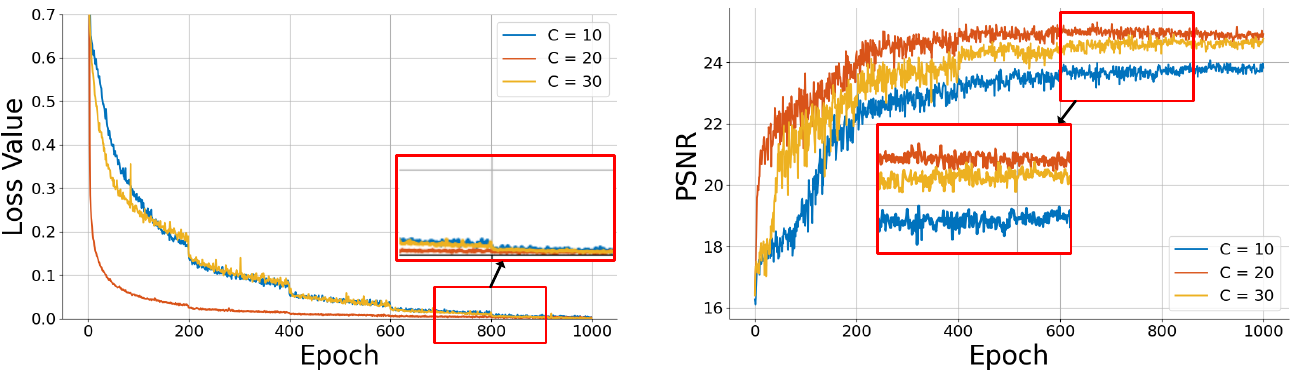}
		\caption{Ablation Study on different depths. C represents the initial number of channels expanded after entering the network, and it can be seen from the figure that the network performs best when C = 20.}
		\label{fig16}
		\end{figure}

		In this subsection, we investigate the impact of network depth on performance. Specifically, we set the number of channels expanded after the first convolutional layer to C and experiment with C = 10, 20, 30. For these three configurations, we train the network for 1000 epochs. As shown in Fig. \ref{fig16}, the left graph illustrates the loss decline curve on the validation set, while the right graph shows the PSNR improvement curve.

        When C = 10, although the network has fewer parameters and faster inference speed, it is too shallow, resulting in slow fitting speed and poor learning quality. When C = 30, the network is too deep with more parameters, leading to overfitting and poor performance. Additionally, the network has an excessive number of parameters when C = 30, as shown in Table V. Comparatively, the network achieves optimal performance when C = 20.
	
		\subsubsection{Contribution of multistage loss}

\begin{figure}[!h]
			\centering
			\includegraphics[width=1.0\columnwidth]{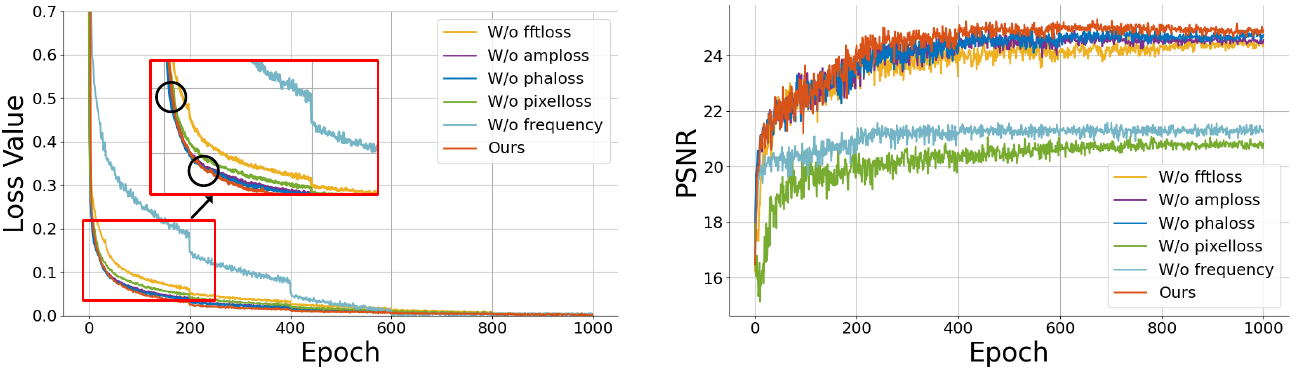}
			\caption{Ablation Study on multistage loss. The left graph shows the loss decrease curves, and the right graph shows the PSNR increase curves. Each experiment was conducted for 1000 epochs.}
			\label{fig17}
		\end{figure}

		In this subsection, we validated the benefits of multi-level loss for the proposed model. As shown in Fig. \ref{fig17}, the left graph depicts the loss decline curves when using different losses, and the right graph illustrates the PSNR increase curves. "W/o frequency" indicates the absence of all frequency domain losses, while "W/o pixelloss" represents the absence of spatial domain loss. The remaining configurations demonstrate the results when each of the three frequency domain losses is individually omitted. From the enlarged area of the left graph, it can be observed that when only spatial or frequency domain information is used for supervision, the fitting speed is significantly slower. When both spatial and frequency domain information are used for supervision simultaneously, the convergence speed of the model is similar initially, but after 100 epochs, the model with the complete loss converges significantly faster. Notably, in the case of "W/o fftloss," although the model uses the corresponding frequency domain information for supervision in both stages, the lack of overall frequency domain guidance makes it more difficult for the model to fit, and the final performance is worse.
		
		\subsubsection{Compared with transformer blocks}
		Finally, we evaluated the effectiveness and efficiency of the proposed DFFN compared to transformers. Specifically, we replaced the frequency domain branches in DDAB and DDPB with the most advanced transformer blocks \cite{Chen_2023_CVPR} currently available, and only used spatial domain information for supervision, with all other operations being the same as in DFFN. 
		
		\begin{figure}[!h]
			\centering
			\includegraphics[width=1.0\columnwidth]{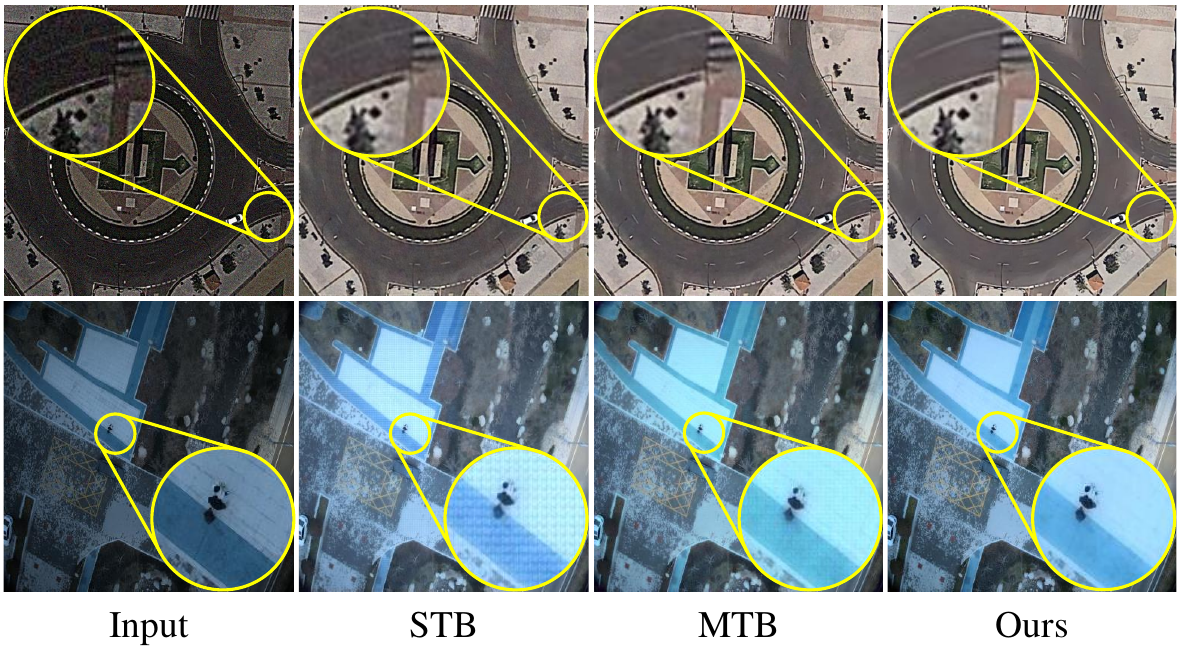}
			\caption{The visualization results compared with transformer blocks, Single Transformer Block (STB) indicates that each block uses only one self-attention block and one FFN, while Multiple Transformer Block (MTB) signifies that multiple self-attention blocks and FFNs are stacked within each transformer block.}
			\label{fig14}
		\end{figure}

        Initially, to ensure fairness, we used a single self-attention block and a single FFN operation in each block, setting num\_head to 2. We refer to this as the Single Transformer Block (STB). Although Transformers are designed to capture long-term dependencies, shallower transformer blocks may struggle to learn rich feature representations, resulting in inferior performance compared to frequency domain networks when processing large images. As shown in Table \ref{tab5}, using only a single operation in transformer blocks leads to suboptimal results while introducing a significant number of parameters.

        Subsequently, we followed the mainstream settings, stacking self-attention blocks and FFN operations in each stage as [2,4,6,8], with num\_head set to [2,2,4,4], which we refer to as Multiple Transformer Block (MTB). However, this setup results in nearly 27M parameters, and the final enhancement effect is still slightly inferior to our method. Additionally, we measured the FLOPs and FPS on 1080p images for STB and our method (MTB could not run due to excessive computational requirements). In comparison, our frequency domain strategy has only 2.57M parameters, with 501 GFLOPs and 4 FPS, achieving the best results. Additionally, as shown in Fig. \ref{fig14}, the visualization demonstrates that our method achieves the best visual effects.

	\section{Conclusion}
		
	In this paper, we propose a two-stage DFFN for enhancing low-light remote sensing images. The DFFN combines the advantages of spatial and frequency domains, consisting of two sub-networks: the amplitude illumination stage and the phase refinement stage. To better integrate dual-domain information, we design the corresponding DDAB \& DDRB. Due to the lack of information interaction in the two-stage network, we further design IFAM to interactively fuse features of different domains, stages, and sizes in the two stages, enhancing the model's contextual representation capability. Furthermore, we introduce the iSAID-dark and darkrs datasets for low-light remote sensing image enhancement.  
		
	Although our method demonstrates good performance in most cases, it still has certain limitations. The presence of numerous FFT and iFFT processes in the network lowers the model's inference speed. Additionally, when there is a significant amount of noise in the environment, DFFN struggles to remove all the noise, which is a common challenge faced by current mainstream LLIE methods. In the future, we will focus on addressing the denoising problem in extremely low-light environments while improving the model's inference speed, so that our method can be better applied to real-world scenarios.

\bibliographystyle{IEEEtran}
\bibliography{DFFN}

\begin{thebibliography}{10}
\providecommand{\url}[1]{#1}
\csname url@samestyle\endcsname
\providecommand{\newblock}{\relax}
\providecommand{\bibinfo}[2]{#2}
\providecommand{\BIBentrySTDinterwordspacing}{\spaceskip=0pt\relax}
\providecommand{\BIBentryALTinterwordstretchfactor}{4}
\providecommand{\BIBentryALTinterwordspacing}{\spaceskip=\fontdimen2\font plus
\BIBentryALTinterwordstretchfactor\fontdimen3\font minus
  \fontdimen4\font\relax}
\providecommand{\BIBforeignlanguage}[2]{{%
\expandafter\ifx\csname l@#1\endcsname\relax
\typeout{** WARNING: IEEEtran.bst: No hyphenation pattern has been}%
\typeout{** loaded for the language `#1'. Using the pattern for}%
\typeout{** the default language instead.}%
\else
\language=\csname l@#1\endcsname
\fi
#2}}
\providecommand{\BIBdecl}{\relax}
\BIBdecl

\bibitem{10175627}
J.~Zhang, J.~Lei, W.~Xie, Y.~Li, G.~Yang, and X.~Jia, ``Guided hybrid
  quantization for object detection in remote sensing imagery via one-to-one
  self-teaching,'' \emph{IEEE Transactions on Geoscience and Remote Sensing},
  vol.~61, pp. 1--15, 2023.

\bibitem{zhang2021rethinking}
Y.-f. Zhang, J.~Zheng, L.~Li, N.~Liu, W.~Jia, X.~Fan, C.~Xu, and X.~He,
  ``Rethinking feature aggregation for deep rgb-d salient object detection,''
  \emph{Neurocomputing}, vol. 423, pp. 463--473, 2021.

\bibitem{9743313}
L.~Ma, R.~Liu, Y.~Wang, X.~Fan, and Z.~Luo, ``Low-light image enhancement via
  self-reinforced retinex projection model,'' \emph{IEEE Transactions on
  Multimedia}, vol.~25, pp. 3573--3586, 2023.

\bibitem{li2021low}
J.~Li, X.~Feng, and Z.~Hua, ``Low-light image enhancement via
  progressive-recursive network,'' \emph{IEEE Transactions on Circuits and
  Systems for Video Technology}, vol.~31, no.~11, pp. 4227--4240, 2021.

\bibitem{zhang2021better}
Y.~Zhang, X.~Di, B.~Zhang, R.~Ji, and C.~Wang, ``Better than reference in
  low-light image enhancement: conditional re-enhancement network,'' \emph{IEEE
  Transactions on Image Processing}, vol.~31, pp. 759--772, 2021.

\bibitem{10219916}
Y.~Huang, X.~Tu, G.~Fu, T.~Liu, B.~Liu, M.~Yang, and Z.~Feng, ``Low-light image
  enhancement by learning contrastive representations in spatial and frequency
  domains,'' in \emph{2023 IEEE International Conference on Multimedia and
  Expo}, 2023, pp. 1307--1312.

\bibitem{fan2022multiscale}
G.-D. Fan, B.~Fan, M.~Gan, G.-Y. Chen, and C.~P. Chen, ``Multiscale low-light
  image enhancement network with illumination constraint,'' \emph{IEEE
  Transactions on Circuits and Systems for Video Technology}, vol.~32, no.~11,
  pp. 7403--7417, 2022.

\bibitem{ma2021learning}
L.~Ma, R.~Liu, J.~Zhang, X.~Fan, and Z.~Luo, ``Learning deep context-sensitive
  decomposition for low-light image enhancement,'' \emph{IEEE Transactions on
  Neural Networks and Learning Systems}, vol.~33, no.~10, pp. 5666--5680, 2021.

\bibitem{li2023low}
Z.~Li, Y.~Wang, and J.~Zhang, ``Low-light image enhancement with knowledge
  distillation,'' \emph{Neurocomputing}, vol. 518, pp. 332--343, 2023.

\bibitem{9812717}
G.~Fan, M.~Gan, B.~Fan, and C.~L.~P. Chen, ``Multiscale cross-connected
  dehazing network with scene depth fusion,'' \emph{IEEE Transactions on Neural
  Networks and Learning Systems}, vol.~35, no.~2, pp. 1598--1612, 2024.

\bibitem{zhou2023underwater}
J.~Zhou, Q.~Liu, Q.~Jiang, W.~Ren, K.-M. Lam, and W.~Zhang, ``Underwater
  camera: Improving visual perception via adaptive dark pixel prior and color
  correction,'' \emph{International Journal of Computer Vision}, pp. 1--19,
  2023.

\bibitem{10234460}
S.~Yuan, L.~Li, H.~Chen, and X.~Li, ``Surface defect detection of highly
  reflective leather based on dual-mask-guided deep-learning model,''
  \emph{IEEE Transactions on Instrumentation and Measurement}, vol.~72, pp.
  1--13, 2023.

\bibitem{xu2022snr}
X.~Xu, R.~Wang, C.-W. Fu, and J.~Jia, ``Snr-aware low-light image
  enhancement,'' in \emph{Proceedings of the IEEE/CVF Conference on Computer
  Vision and Pattern Recognition}, 2022, pp. 17\,714--17\,724.

\bibitem{mittal2012making}
A.~Mittal, R.~Soundararajan, and A.~C. Bovik, ``Making a “completely blind”
  image quality analyzer,'' \emph{IEEE Signal processing letters}, vol.~20,
  no.~3, pp. 209--212, 2012.

\bibitem{wang2004image}
Z.~Wang, A.~C. Bovik, H.~R. Sheikh, and E.~P. Simoncelli, ``Image quality
  assessment: from error visibility to structural similarity,'' \emph{IEEE
  Transactions on Image Processing}, vol.~13, no.~4, pp. 600--612, 2004.

\bibitem{huynh2008scope}
Q.~Huynh-Thu and M.~Ghanbari, ``Scope of validity of psnr in image/video
  quality assessment,'' \emph{Electronics letters}, vol.~44, no.~13, pp.
  800--801, 2008.

\bibitem{9992208}
J.-N. Su, M.~Gan, G.-Y. Chen, J.-L. Yin, and C.~L.~P. Chen, ``Global learnable
  attention for single image super-resolution,'' \emph{IEEE Transactions on
  Pattern Analysis and Machine Intelligence}, vol.~45, no.~7, pp. 8453--8465,
  2023.

\bibitem{10196016}
S.~Zhang, J.~Zhang, X.~Wang, J.~Wang, and Z.~Wu, ``Els2t: Efficient lightweight
  spectral–spatial transformer for hyperspectral image classification,''
  \emph{IEEE Transactions on Geoscience and Remote Sensing}, vol.~61, pp.
  1--16, 2023.

\bibitem{10463068}
J.~Feng, Q.~Wang, G.~Zhang, X.~Jia, and J.~Yin, ``Cat: Center attention
  transformer with stratified spatial-spectral token for hyperspectral image
  classification,'' \emph{IEEE Transactions on Geoscience and Remote Sensing},
  pp. 1--1, 2024.

\bibitem{10167502}
C.~Zhao, B.~Qin, S.~Feng, W.~Zhu, W.~Sun, W.~Li, and X.~Jia, ``Hyperspectral
  image classification with multi-attention transformer and adaptive superpixel
  segmentation-based active learning,'' \emph{IEEE Transactions on Image
  Processing}, vol.~32, pp. 3606--3621, 2023.

\bibitem{10382425}
J.-N. Su, M.~Gan, G.-Y. Chen, W.~Guo, and C.~L.~P. Chen, ``High-similarity-pass
  attention for single image super-resolution,'' \emph{IEEE Transactions on
  Image Processing}, vol.~33, pp. 610--624, 2024.

\bibitem{dosovitskiy2020image}
A.~Dosovitskiy, L.~Beyer, A.~Kolesnikov, D.~Weissenborn, X.~Zhai,
  T.~Unterthiner, M.~Dehghani, M.~Minderer, G.~Heigold, S.~Gelly \emph{et~al.},
  ``An image is worth 16x16 words: Transformers for image recognition at
  scale,'' \emph{arXiv preprint arXiv:2010.11929}, 2020.

\bibitem{10298274}
W.~Dong, Y.~Yang, J.~Qu, Y.~Li, Y.~Yang, and X.~Jia, ``Feature pyramid fusion
  network for hyperspectral pansharpening,'' \emph{IEEE Transactions on Neural
  Networks and Learning Systems}, pp. 1--13, 2023.

\bibitem{10177702}
J.~Zhou, B.~Li, D.~Zhang, J.~Yuan, W.~Zhang, Z.~Cai, and J.~Shi, ``Ugif-net: An
  efficient fully guided information flow network for underwater image
  enhancement,'' \emph{IEEE Transactions on Geoscience and Remote Sensing},
  vol.~61, pp. 1--17, 2023.

\bibitem{ma2023bilevel}
L.~Ma, D.~Jin, N.~An, J.~Liu, X.~Fan, Z.~Luo, and R.~Liu, ``Bilevel fast scene
  adaptation for low-light image enhancement,'' \emph{International Journal of
  Computer Vision}, pp. 1--19, 2023.

\bibitem{xing2023clegan}
L.~Xing, H.~Qu, S.~Xu, and Y.~Tian, ``Clegan: Towards low-light image
  enhancement for uavs via self-similarity exploitation,'' \emph{IEEE
  Transactions on Geoscience and Remote Sensing}, 2023.

\bibitem{singh2022low}
A.~Singh, A.~Chougule, P.~Narang, V.~Chamola, and F.~R. Yu, ``Low-light image
  enhancement for uavs with multi-feature fusion deep neural networks,''
  \emph{IEEE Geoscience and Remote Sensing Letters}, vol.~19, pp. 1--5, 2022.

\bibitem{wang2023fourllie}
C.~Wang, H.~Wu, and Z.~Jin, ``Fourllie: Boosting low-light image enhancement by
  fourier frequency information,'' in \emph{Proceedings of the 31st ACM
  International Conference on Multimedia}, 2023, pp. 7459--7469.

\bibitem{oppenheim1979phase}
A.~Oppenheim, J.~Lim, G.~Kopec, and S.~Pohlig, ``Phase in speech and
  pictures,'' in \emph{ICASSP'79. IEEE International Conference on Acoustics,
  Speech, and Signal Processing}, vol.~4.\hskip 1em plus 0.5em minus
  0.4em\relax IEEE, 1979, pp. 632--637.

\bibitem{li2023embedding}
C.~Li, C.-L. Guo, M.~Zhou, Z.~Liang, S.~Zhou, R.~Feng, and C.~C. Loy,
  ``Embedding fourier for ultra-high-definition low-light image enhancement,''
  \emph{arXiv preprint arXiv:2302.11831}, 2023.

\bibitem{zhu2020eemefn}
M.~Zhu, P.~Pan, W.~Chen, and Y.~Yang, ``Eemefn: Low-light image enhancement via
  edge-enhanced multi-exposure fusion network,'' in \emph{Proceedings of the
  AAAI conference on artificial intelligence}, vol.~34, no.~07, 2020, pp.
  13\,106--13\,113.

\bibitem{mou2022deep}
C.~Mou, Q.~Wang, and J.~Zhang, ``Deep generalized unfolding networks for image
  restoration,'' in \emph{Proceedings of the IEEE/CVF Conference on Computer
  Vision and Pattern Recognition}, 2022, pp. 17\,399--17\,410.

\bibitem{arici2009histogram}
T.~Arici, S.~Dikbas, and Y.~Altunbasak, ``A histogram modification framework
  and its application for image contrast enhancement,'' \emph{IEEE Transactions
  on Image Processing}, vol.~18, no.~9, pp. 1921--1935, 2009.

\bibitem{celik2011contextual}
T.~Celik and T.~Tjahjadi, ``Contextual and variational contrast enhancement,''
  \emph{IEEE Transactions on Image Processing}, vol.~20, no.~12, pp.
  3431--3441, 2011.

\bibitem{ibrahim2007brightness}
H.~Ibrahim and N.~S.~P. Kong, ``Brightness preserving dynamic histogram
  equalization for image contrast enhancement,'' \emph{IEEE Transactions on
  Consumer Electronics}, vol.~53, no.~4, pp. 1752--1758, 2007.

\bibitem{huang2012efficient}
S.-C. Huang, F.-C. Cheng, and Y.-S. Chiu, ``Efficient contrast enhancement
  using adaptive gamma correction with weighting distribution,'' \emph{IEEE
  Transactions on Image Processing}, vol.~22, no.~3, pp. 1032--1041, 2012.

\bibitem{singh2017novel}
H.~Singh, A.~Kumar, L.~Balyan, and G.~K. Singh, ``A novel optimally gamma
  corrected intensity span maximization approach for dark image enhancement,''
  in \emph{2017 22nd International Conference on Digital Signal
  Processing}.\hskip 1em plus 0.5em minus 0.4em\relax IEEE, 2017, pp. 1--5.

\bibitem{fu2016weighted}
X.~Fu, D.~Zeng, Y.~Huang, X.-P. Zhang, and X.~Ding, ``A weighted variational
  model for simultaneous reflectance and illumination estimation,'' in
  \emph{Proceedings of the IEEE conference on computer vision and pattern
  recognition}, 2016, pp. 2782--2790.

\bibitem{jobson1997multiscale}
D.~J. Jobson, Z.-u. Rahman, and G.~A. Woodell, ``A multiscale retinex for
  bridging the gap between color images and the human observation of scenes,''
  \emph{IEEE Transactions on Image Processing}, vol.~6, no.~7, pp. 965--976,
  1997.

\bibitem{xu2024seeing}
C.~Xu, H.~Fu, L.~Ma, W.~Jia, C.~Zhang, F.~Xia, X.~Ai, B.~Li, and W.~Zhang,
  ``Seeing text in the dark: Algorithm and benchmark,'' \emph{arXiv preprint
  arXiv:2404.08965}, 2024.

\bibitem{li2018structure}
M.~Li, J.~Liu, W.~Yang, X.~Sun, and Z.~Guo, ``Structure-revealing low-light
  image enhancement via robust retinex model,'' \emph{IEEE Transactions on
  Image Processing}, vol.~27, no.~6, pp. 2828--2841, 2018.

\bibitem{guo2016lime}
X.~Guo, Y.~Li, and H.~Ling, ``Lime: Low-light image enhancement via
  illumination map estimation,'' \emph{IEEE Transactions on Image Processing},
  vol.~26, no.~2, pp. 982--993, 2016.

\bibitem{shen2023mutual}
H.~Shen, Z.-Q. Zhao, Y.~Zhang, and Z.~Zhang, ``Mutual information-driven triple
  interaction network for efficient image dehazing,'' in \emph{Proceedings of
  the 31st ACM International Conference on Multimedia}, 2023, pp. 7--16.

\bibitem{liang2022self}
J.~Liang, Y.~Xu, Y.~Quan, B.~Shi, and H.~Ji, ``Self-supervised low-light image
  enhancement using discrepant untrained network priors,'' \emph{IEEE
  Transactions on Circuits and Systems for Video Technology}, vol.~32, no.~11,
  pp. 7332--7345, 2022.

\bibitem{10147801}
Y.~Luo, B.~You, G.~Yue, and J.~Ling, ``Pseudo-supervised low-light image
  enhancement with mutual learning,'' \emph{IEEE Transactions on Circuits and
  Systems for Video Technology}, vol.~34, no.~1, pp. 85--96, 2024.

\bibitem{ma2022learning}
L.~Ma, R.~Liu, J.~Zhang, X.~Fan, and Z.~Luo, ``Learning deep context-sensitive
  decomposition for low-light image enhancement.'' \emph{IEEE Transactions on
  Neural Networks and Learning Systems}, vol.~33, no.~10, pp. 5666--5680, 2022.

\bibitem{li2022learning}
C.~Li, C.~Guo, and C.~C. Loy, ``Learning to enhance low-light image via
  zero-reference deep curve estimation,'' \emph{IEEE Transactions on Pattern
  Analysis and Machine Intelligence}, vol.~44, no.~8, pp. 4225--4238, 2022.

\bibitem{wang2023low}
Y.~Wang, Z.~Liu, J.~Liu, S.~Xu, and S.~Liu, ``Low-light image enhancement with
  illumination-aware gamma correction and complete image modelling network,''
  in \emph{Proceedings of the IEEE/CVF International Conference on Computer
  Vision}, 2023, pp. 13\,128--13\,137.

\bibitem{10136217}
L.~Xing, H.~Qu, S.~Xu, and Y.~Tian, ``Clegan: Toward low-light image
  enhancement for uavs via self-similarity exploitation,'' \emph{IEEE
  Transactions on Geoscience and Remote Sensing}, vol.~61, pp. 1--14, 2023.

\bibitem{huang2022deep}
J.~Huang, Y.~Liu, F.~Zhao, K.~Yan, J.~Zhang, Y.~Huang, M.~Zhou, and Z.~Xiong,
  ``Deep fourier-based exposure correction network with spatial-frequency
  interaction,'' in \emph{European Conference on Computer Vision}.\hskip 1em
  plus 0.5em minus 0.4em\relax Springer, 2022, pp. 163--180.

\bibitem{mao2021deep}
X.~Mao, Y.~Liu, W.~Shen, Q.~Li, and Y.~Wang, ``Deep residual fourier
  transformation for single image deblurring,'' \emph{arXiv preprint
  arXiv:2111.11745}, vol.~2, no.~3, p.~5, 2021.

\bibitem{suvorov2022resolution}
R.~Suvorov, E.~Logacheva, A.~Mashikhin, A.~Remizova, A.~Ashukha, A.~Silvestrov,
  N.~Kong, H.~Goka, K.~Park, and V.~Lempitsky, ``Resolution-robust large mask
  inpainting with fourier convolutions,'' in \emph{Proceedings of the IEEE/CVF
  Winter Conference on Applications of Computer Vision}, 2022, pp. 2149--2159.

\bibitem{fuoli2021fourier}
D.~Fuoli, L.~Van~Gool, and R.~Timofte, ``Fourier space losses for efficient
  perceptual image super-resolution,'' in \emph{Proceedings of the IEEE/CVF
  International Conference on Computer Vision}, 2021, pp. 2360--2369.

\bibitem{zheng2022hfa}
H.~Zheng, M.~Gong, T.~Liu, F.~Jiang, T.~Zhan, D.~Lu, and M.~Zhang, ``Hfa-net:
  High frequency attention siamese network for building change detection in vhr
  remote sensing images,'' \emph{Pattern Recognition}, vol. 129, p. 108717,
  2022.

\bibitem{zhou2022spatial}
M.~Zhou, J.~Huang, K.~Yan, H.~Yu, X.~Fu, A.~Liu, X.~Wei, and F.~Zhao,
  ``Spatial-frequency domain information integration for pan-sharpening,'' in
  \emph{European Conference on Computer Vision}.\hskip 1em plus 0.5em minus
  0.4em\relax Springer, 2022, pp. 274--291.

\bibitem{yu2022snnfd}
B.~Yu, A.~Yang, F.~Chen, N.~Wang, and L.~Wang, ``Snnfd, spiking neural
  segmentation network in frequency domain using high spatial resolution images
  for building extraction,'' \emph{International Journal of Applied Earth
  Observation and Geoinformation}, vol. 112, p. 102930, 2022.

\bibitem{zamir2021multi}
S.~W. Zamir, A.~Arora, S.~Khan, M.~Hayat, F.~S. Khan, M.-H. Yang, and L.~Shao,
  ``Multi-stage progressive image restoration,'' in \emph{Proceedings of the
  IEEE/CVF Conference on Computer Vision and Pattern Recognition}, 2021, pp.
  14\,821--14\,831.

\bibitem{ma2022toward}
L.~Ma, T.~Ma, R.~Liu, X.~Fan, and Z.~Luo, ``Toward fast, flexible, and robust
  low-light image enhancement,'' in \emph{Proceedings of the IEEE/CVF
  Conference on Computer Vision and Pattern Recognition}, 2022, pp. 5637--5646.

\bibitem{wang2023ultra}
T.~Wang, K.~Zhang, T.~Shen, W.~Luo, B.~Stenger, and T.~Lu,
  ``Ultra-high-definition low-light image enhancement: A benchmark and
  transformer-based method,'' in \emph{Proceedings of the AAAI Conference on
  Artificial Intelligence}, vol.~37, no.~3, 2023, pp. 2654--2662.

\bibitem{zheng2023empowering}
N.~Zheng, M.~Zhou, Y.~Dong, X.~Rui, J.~Huang, C.~Li, and F.~Zhao, ``Empowering
  low-light image enhancer through customized learnable priors,'' in
  \emph{Proceedings of the IEEE/CVF International Conference on Computer
  Vision}, 2023, pp. 12\,559--12\,569.

\bibitem{yang2023learning}
K.-F. Yang, C.~Cheng, S.-X. Zhao, H.-M. Yan, X.-S. Zhang, and Y.-J. Li,
  ``Learning to adapt to light,'' \emph{International Journal of Computer
  Vision}, vol. 131, no.~4, pp. 1022--1041, 2023.

\bibitem{yang2023implicit}
S.~Yang, M.~Ding, Y.~Wu, Z.~Li, and J.~Zhang, ``Implicit neural representation
  for cooperative low-light image enhancement,'' in \emph{Proceedings of the
  IEEE/CVF International Conference on Computer Vision}, 2023, pp.
  12\,918--12\,927.

\bibitem{waqas2019isaid}
S.~Waqas~Zamir, A.~Arora, A.~Gupta, S.~Khan, G.~Sun, F.~Shahbaz~Khan, F.~Zhu,
  L.~Shao, G.-S. Xia, and X.~Bai, ``isaid: A large-scale dataset for instance
  segmentation in aerial images,'' in \emph{Proceedings of the IEEE/CVF
  Conference on Computer Vision and Pattern Recognition Workshops}, 2019, pp.
  28--37.

\bibitem{wei2018deep}
C.~Wei, W.~Wang, W.~Yang, and J.~Liu, ``Deep retinex decomposition for
  low-light enhancement,'' \emph{British Machine Vision Conference}, 2018.

\bibitem{zhang2018unreasonable}
R.~Zhang, P.~Isola, A.~A. Efros, E.~Shechtman, and O.~Wang, ``The unreasonable
  effectiveness of deep features as a perceptual metric,'' in \emph{Proceedings
  of the IEEE Conference on Computer Vision and pattern recognition}, 2018, pp.
  586--595.

\bibitem{Chen_2023_CVPR}
X.~Chen, H.~Li, M.~Li, and J.~Pan, ``Learning a sparse transformer network for
  effective image deraining,'' in \emph{Proceedings of the IEEE/CVF Conference
  on Computer Vision and Pattern Recognition}, June 2023, pp. 5896--5905.

\end{thebibliography}
\vspace{-35px}
\begin{IEEEbiography}[{\includegraphics[width=1in,height=1.25in,clip,keepaspectratio]{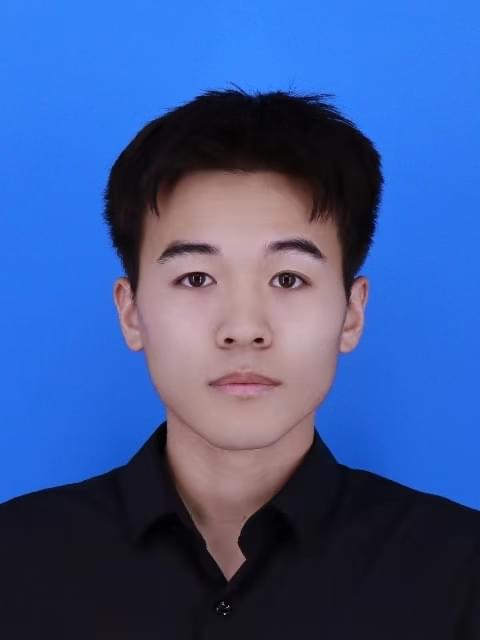}}]{Zishu Yao}received the B.S. degree from Shandong University of Finance and Economics, China, in 2021, where he is currently pursuing the M.S. degree with the Qingdao University. His research interests are in image processing, machine learning, and computer vision.	
\end{IEEEbiography}
\vspace{-25px}

\begin{IEEEbiography}[{\includegraphics[width=1in,height=1.25in,clip,keepaspectratio]{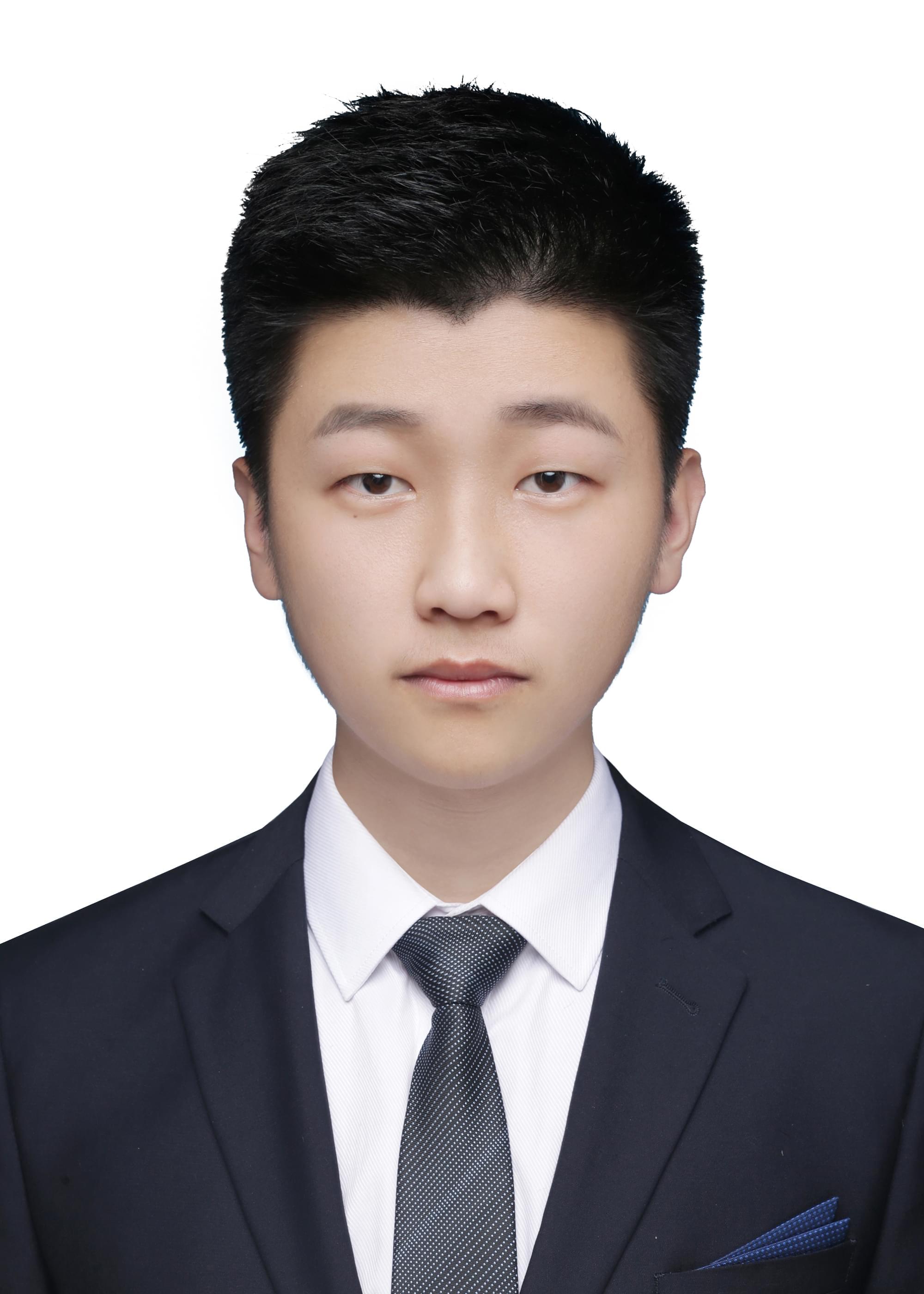}}]{Guodong Fan}
	received the M. Eng. degree from Shandong Technology and Business University, China, in 2021, where he is currently pursuing the Ph.D. degree with the Qingdao University.  His research interests are in image processing, machine learning, and computer vision.
\end{IEEEbiography}
\vspace{-125px}

\begin{IEEEbiography}[{\includegraphics[width=1in,height=1.25in,clip,keepaspectratio]{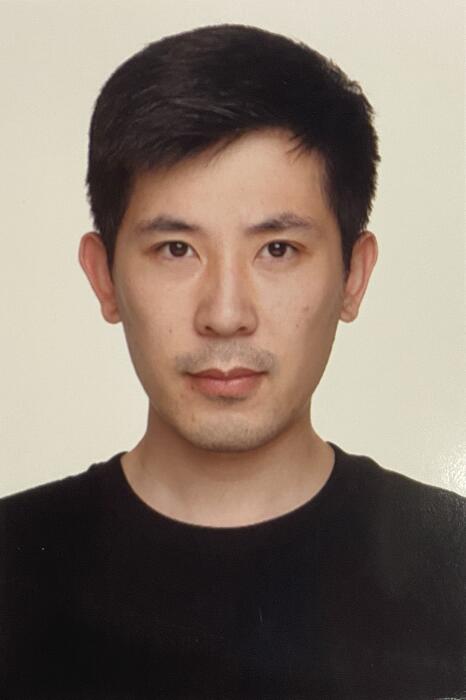}}]{Jinfu Fan} the Ph.D. degree in control theory and control engineering from Tongji University, Shanghai. He is currently an associate professor with the College of Computer Science and Technology, Qingdao University, China. His current research interests include machine learning and computer vision, image processing, graph network, especially in learning from partial label data.
\end{IEEEbiography} 
\vspace{-125px}

\begin{IEEEbiography}[{\includegraphics[width=1in,height=1.25in,clip,keepaspectratio]{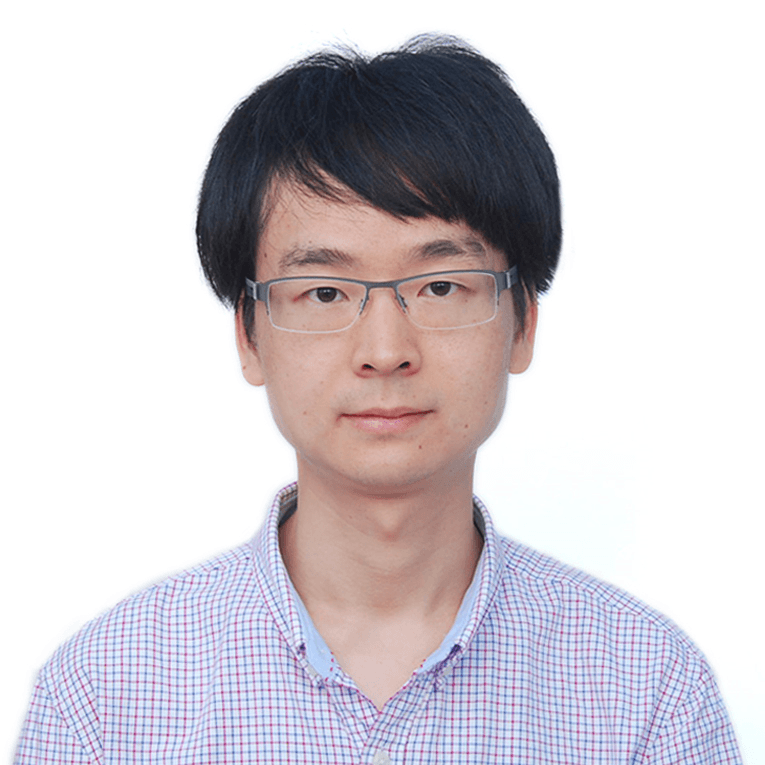}}]{Min Gan}(Senior Member,~IEEE)
	received the B. S. degree in Computer Science and Engineering from Hubei University of Technology, Wuhan, China, in 2004, and the Ph.D. degree in Control Science and Engineering from Central South University, Changsha, China, in 2010.  His current research interests include machine learning, inverse problems, and image processing.
\end{IEEEbiography} 
\vspace{-125px}

\begin{IEEEbiography}[{\includegraphics[width=1in,height=1.25in,clip,keepaspectratio]{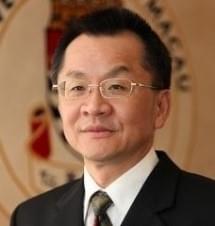}}]{C. L. Philip Chen} (Fellow,~IEEE) is the Chair Professor and Dean of the College of Computer Science and Engineering, South China University of Technology. Being a Program Evaluator of the Accreditation Board of Engineering and Technology Education(ABET) in the U.S., for computer engineering, electrical engineering, and software engineering programs, he successfully architects the University of Macau's Engineering and Computer Science programs receiving accreditations from Washington/Seoul Accord through Hong Kong Institute of Engineers(HKIE), of which is considered as his utmost contribution in engineering/computer Science education for Macau as the former Dean of the Faculty of Science and Technology. 
	Prof. Chen is a Fellow of IEEE, AAAS, IAPR, CAA, and HKIE; a member of Academia Europaea(AE), European Academy of Sciences and Arts(EASA), and International Academy of Systems and Cybernetics Science(IASCYS). He received IEEE Norbert Wiener Award in 2018 for his contribution in systems and cybernetics, and machine learnings.  He is also a Highly Cited Researcher by Clarivate Analytics in 2018 and 2021.
\end{IEEEbiography}

\end{document}